\documentclass[runningheads]{llncs}
\usepackage{graphicx}
\usepackage{hyperref}
\usepackage{tikz}
\usepackage{comment}
\usepackage{amsmath,amssymb} % define this before the line numbering.
\usepackage{color}
\usepackage{xspace}
\usepackage[accsupp]{axessibility}  % Improves PDF readability for those with disabilities.

\newcommand{\marcus}[1]{{\textcolor{orange}{(Marcus: #1)}}}

\newcommand{\ls}[1]{}
\newcommand{\sh}[1]{}

\newcommand{\Method}{Learn2Augment\xspace}
\newcommand{\MethodShort}{L2A\xspace}

\newcommand{\SemanticMatch}{Semantic Matching\xspace}
\newcommand{\SemanticMatchShort}{SM\xspace}

\newcommand{\VideoMix}{Video Compositing\xspace}

\begin{document}
% \linenumbers
\pagestyle{headings}
\mainmatter
\def\ECCVSubNumber{5234}  % Insert your submission number here

\title{Learn2Augment: Learning to Composite Videos for Data Augmentation in Action Recognition} % Replace with your title

\begin{comment}
\titlerunning{ECCV-22 submission ID \ECCVSubNumber} 
\authorrunning{ECCV-22 submission ID \ECCVSubNumber} 
\author{Anonymous ECCV submission}
\institute{Paper ID \ECCVSubNumber}
\end{comment}
%******************

%\begin{comment}
\titlerunning{Learn2Augment}
\author{Shreyank N Gowda\inst{1} \and
Marcus Rohrbach\inst{2} \and \\
Frank Keller\inst{1}  \and
Laura Sevilla-Lara\inst{1}}
\authorrunning{Shreyank N Gowda et al.}
\institute{University of Edinburgh \and
Meta AI\\
}
%end{comment}
%******************

\maketitle

\begin{abstract}
We address the problem of data augmentation for video action recognition. Standard augmentation strategies in video are hand-designed and sample the space of possible augmented data points either at random, without knowing which augmented points will be better, or through heuristics. We propose to learn what makes a ``good'' video for action recognition and select only high-quality samples for augmentation. In particular, we choose video compositing of a foreground and a background video as the data augmentation process, which results in diverse and realistic new samples. We learn which pairs of videos to augment {\em without} having to actually composite them. This reduces the space of possible augmentations, which has two advantages: it saves computational cost and increases the accuracy of the final trained classifier, as the augmented pairs are of higher quality than average. We present experimental results on the entire spectrum of training settings: few-shot, semi-supervised and fully supervised. We observe consistent improvements across all of them over prior work and baselines on Kinetics, UCF101, HMDB51, and achieve a new state-of-the-art on settings with limited data. We see improvements of up to 8.6\% in the semi-supervised setting. Project Page: \url{https://sites.google.com/view/learn2augment/home}
%\marcus{would it be better to push the improvements here} Code and models will be made available.\marcus{I find this a strange sentence in the abstract.}  
%\keywords{We would like to encourage you to list your keywords within the abstract section}
\end{abstract}

\begin{figure}
\begin{center}
    \centering
    %\captionsetup{type=figure}
    \includegraphics[width=0.99\textwidth]{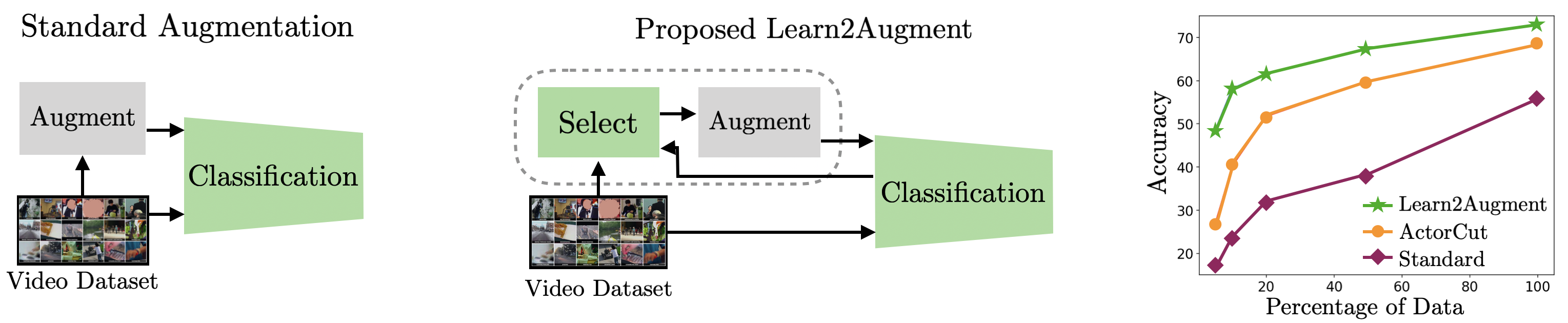}
    \caption{Standard video augmentation techniques generate data using hand-designed heuristics (left). We propose to learn to select videos for augmentation, based on how effective they will be for learning to classify (middle). Our approach, \Method, improves classification across datasets and settings, including UCF101 (right). }%, up to 6.2\% (c). \marcus{why don't we have 100\% in this figure (c)? Is that not what we have in Table 6?}
    %Comparison of accuracy on different percentages of labeled data using UCF101 dataset. We compare against a supervised baseline and recent state-of-the-art TCL \cite{tcl}. We see that we are consistently outperforming TCL on every setting and with a difference of up to 6.2\%
    %}
    %\marcus{add Learn2Augment to fig}
\end{center}
\end{figure}

\section{Introduction}
\label{sec:intro}

%\ls{mention we are sota}
Large-scale datasets have played a key role in the progress of research across AI problems. In computer vision, neural networks have existed for decades, but one of the enabling factors for the current revolution was the development of the large ImageNet~\cite{deng2009imagenet}.  
In the video domain, manually collecting and annotating data can be a prohibitively expensive process. In video action recognition, for example, collecting data requires an immense amount of manual labor, as it involves finding suitable videos, trimming them and classifying them. 

% \begin{figure*}[t]
%     \centering
%     \includegraphics[width=0.45\textwidth]{./figures/teaser_standard_proposed.png}
%     \hspace{1.2cm}
%     \includegraphics[width=0.45\textwidth]{./figures/teaser.png}
%     \caption{Comparison of accuracy on different percentages of labeled data using UCF101 dataset. We compare against a supervised baseline and recent state-of-the-art TCL \cite{tcl}. We see that we are consistently outperforming TCL on every setting and with a difference of upto 6.2\%. \ls{can you try to plot all the methods from Table 3 and see what it looks like? Also, perhaps remove the ``supervised'', which I assume is disregarding unlabeled, which is not something we're actually introducing. }}
%     \label{fig:teaser}
% \end{figure*}

Recent efforts in video focus on relieving the strong dependency of current methods to the size of labeled datasets. %has been the focus of many research efforts in the last few years. 
% In the last few years, research efforts have turned to ways of relieving the dependency on labeled data. 
%These efforts include learning from semi-supervised data weakly labeled, few-shot and zero-shot and data augmentation. 
Some of these efforts \cite{sun2021autoflow,actorcut} involve increasing the number of data samples through data augmentation. This strategy aims to create new videos in the training set by performing transformations on the original annotated videos, where labels are known. This process adds diversity to the training data, while new videos are still realistic and plausible. In the simplest version of data augmentation in video, new data samples are generated by flipping the input video horizontally, or by cropping a subsection of the video. New methods~\cite{actorcut,videomix} propose more sophisticated processes like combining two videos. VideoMix~\cite{videomix}
randomly crops regions of one video and pastes them onto another. ActorCut~\cite{actorcut} goes one step further and uses the bounding box detections of humans on one video to paste them onto the background of another video. This increases the diversity of the new videos, and despite the lack of visual realism of the resulting videos, this strategy helps.   

However, as datasets become larger, such data augmentation strategies become computationally expensive. % and not all combinations of existing videos yield high quality novel videos.
%\marcus{I think this is only true for composites, cropping etc. can be very cheap, especially the latest patch based transformer models}. 
The search space of possible video pairs and transformations is enormous and difficult to explore. The solution is often to sample the space randomly, or to manually design augmentation heuristics. Any exploration process is particularly burdening in the context of video data, where the augmentation process needs to be repeated in every frame, which may be orders of magnitude more expensive than for images. %Even for short videos, this means that the data augmentation process can be orders of magnitude larger than for images. %\marcus{I think this is true for compositing what we and actormix is doing, but not necessarily for others, such cropping, blurring. e.g. you could just mix frames/clips from different videos}.  %~\ls{maybe add something about possible number of combinations...}

In this paper we address the problem of sampling for data augmentation, and propose to learn to select pairs of videos. We show that this reduces the search space of augmented data points by orders of magnitude and improves the final accuracy of the classifier significantly. We leverage two observations. First, not all data points are as useful for classification. This idea has been exploited in the context of frame or clip selection~\cite{smart,korbar2019scsampler,whatmakes}. Second, we can learn to predict which data points will be useful without actually generating them. This is essential, as the space of transformations is huge, and if we needed to create each candidate augmented video, the process would be prohibitively expensive. %We do augmentation only in the videos that will yield good augmented samples. 

%Given this ability of predicting which videos are good for augmentation,

%Given this ability to select which videos yield good augmented samples, we augment only the videos that will produce the most useful samples. %, and use data augmentation for those. %reduce the expensive step of generating new videos and training with them to only those useful ones. 

More concretely, we propose a data augmentation method which we call \Method. The proposed method contains a ``Selector'' network, which predicts a score of how useful a combination of two videos will be, without having to actually composite them. The Selector is trained using the accuracy of the classification as the cue. Since this metric depends on the classifier, it is not differentiable with respect to the Selector's parameters. Therefore we optimize the network using reinforcement learning. Once the Selector network is trained, we use it to choose good pairs of videos, composite them, and train a classification network. In our experiments, for example in the case of the UCF101 dataset, using the Selector reduces the number of augmented videos by 92\% while increasing the classification accuracy. 

In the proposed method, each augmented video is created from a pair of videos using a composition of the segmented foreground of one video, including actor and objects, onto the background of the other video. This process yields diverse and realistic new data samples, which we demonstrate is important for learning. More concretely, results show an improvement of 4.4\% over using a simpler transformation. 

%The learning problem is to predict a score of how useful the resulting composited video will be, given the two input videos, without having to actually composite it. We train a ``Selector'' network to solve this problem, using the accuracy of the classification as the cue. Since this metric depends on the classifier and not on the Selector, it is not differentiable with respect to the Selector. Therefore we optimize the network using reinforcement learning. Once the Selector network is trained, we use it to choose good pairs of videos, composite them, and train a classification network. In our experiments, for example in the case of the UCF101 dataset, using the Selector reduces the number of augmented videos by 92\% while increasing the classification accuracy. 
%We train a network for this using reinforcement learning, and use it to select pairs of video to composite for augmentation. 

The Selector is indeed useful to reduce the number of videos for training the classifier. % The Selector reduces this space to train the classification network. 
However, we also need to reduce the space of possible pairs for training the Selector network itself. For example, the number of possible pairs of videos in video datasets can be in the order of millions for small datasets or billions for large datasets.
%We do so by choosing pairs of videos that could yield compositions more plausible than average. 
For this, we leverage the natural correlation between the occurrence of foreground activities and background scenes~\cite{Choi-NeurIPS-2019}. This is, it is more likely to find someone playing football in a football field than at a restaurant. Instead of sampling at random the pairs of videos to train the Selector on, we sample pairs from classes that are semantically similar. In particular, we use the class names to obtain a semantic embedding, and match each class to their nearest neighbor in this space. %. We do this by sampling the foreground and background videos from classes that are semantically similar. This leverages the natural correlation between the occurrence of foreground activities and background scenes~\cite{Choi-NeurIPS-2019}.
Experiments show that this extremely simple design choice of Semantic Matching reduces the space of possible pairs of videos by several orders of magnitude (from quadratic to linear on the number of videos). This yields better results than choosing pairs at random, which may result in non-plausible scenarios, or choosing pairs from the same class, which may not add as much diversity. %This reduces the search space from ~1.2M possible pairs to ~99K.

%given two input videos we  train a network to predict which pairs will yield a good training sample, using reinforcement learning. We use this network to select good pairs of videos.  and combine foreground and background into a novel video. 

% We observe that this process allows us to reduce the amount of generated videos by an order of magnitude, while improving the classification accuracy by up to 3.1\%~\ls{this number is different from the abstract}. We test this procedure under many different settings including augmenting the entire dataset, augmenting only a portion, and augmenting when there are only few training samples. Experimental results show an improvement across all of them. 

%\ls{summarize all things we want the reviewer to write in the strengths section}

% * not emphasize RL 
% * we use it bc we can't optimize with traditional gradient descent 

In summary, the proposed \textbf{\Method} contains three core components: a Selector that learns to choose good videos to augment, a \SemanticMatch method that improves optimization, and a \VideoMix that composites video pairs for augmentation. Experimental results show that all components contribute to the performance of the system in different ways, and the overall method obtains state-of-the-art in all datasets, and in all settings that involve limited training data. In addition, in the setting which considers the full training set, the proposed data augmentation technique improves upon the baseline on all datasets, including UCF101, HMDB51, and the large-scale Kinetics-400.

%\ls{we haven't really talked about semantic matching or video composition here. }
%\ls{I'm not 100\% happy with this yet. I'd like to give less importance to semantic match and composition... thoughts?}
% In a nutshell, \textbf{\Method} is made up of three core components:
% %To summarize, we propose the following novel contributions:
% \begin{itemize}
% \item \textbf{\RL} Use of reinforcement learning to train a selection network to decide what pairs of videos would be good to generate a new video. This results in a drop of ~92\% at possible pairs of videos to look at, whilst increasing the accuracy by over 3\%.
%     \item \textbf{\SemanticMatch} Use of semantic class matching to reduce the search space for picking foreground and background samples. This reduces the search space from ~1.2M possible pairs to ~99K.
%     \item \textbf{\VideoMix} Removal of objects along with actors from both foreground and background samples before mixing, whilst also doing cleaner segmentation. This results in a 4.4\% increase in comparison to without.
% \end{itemize}    

\section{Related Work}

\begin{figure*}[h!]
    \centering
    \includegraphics[width=0.85\textwidth]{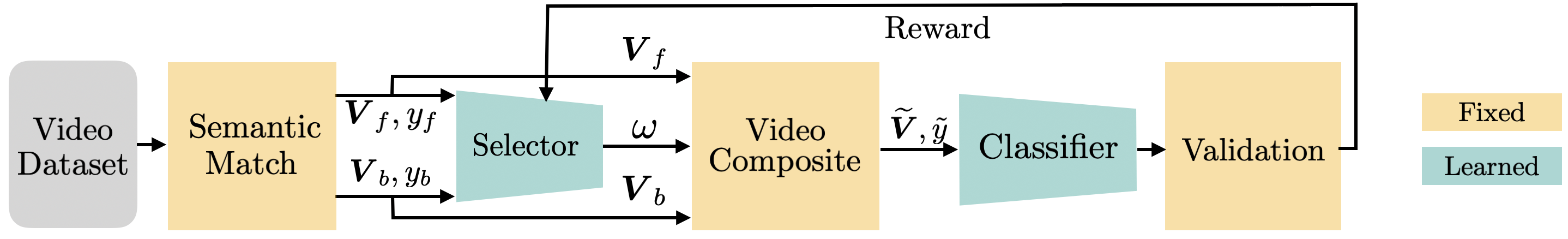}
    %\hspace{0.5cm}
    %\includegraphics[width=0.1\textwidth]{./figures/legend.png} 
    \caption{Overview of the proposed \Method. Given a pair of videos and their labels, a Selector network gives a score $\omega$ of the quality of the potential composited video. At training time, the Selector is trained with the validation loss of the classification network. Once the Selector is trained, pairs of videos are sampled, and only the promising combinations with high score $\omega$ are composited and used for training the classifier. %\marcus{Video Composite change to Video Compositing (VC)}\marcus{Add "(SM)" to Semantic Match}
    %\marcus{This figure misses the \SemanticMatch component}
    %\marcus{I don't understand the yellow "Data generation phase". For me it could just be removed}
    %\marcus{why both $V_1$ and $V_f$, they are the same!? }
    }
    \label{fig:overview}
\end{figure*}
\paragraph{Data Augmentation for Video Action Recognition.}
Standard data augmentation techniques in action recognition include horizontal flip and cropping, where new videos are created by selecting a box at each frame, and then resizing the resulting video to have the same size as the original one. While this strategy helps, generated videos do not add much diversity to the training set. 
Recent efforts such as ActorCut~\cite{actorcut} and VideoMix~\cite{videomix} increase the diversity of new video samples by cutting and pasting the foreground of one video onto another. This general technique of combining two data samples has proven to be quite effective, even in the image domain~\cite{yun2019cutmix}. However, the resulting videos are not very realistic, and are used for training regardless of their quality. Zhang et al.~\cite{aug_gans} go one step further and synthesize new samples using GANs, and use ``self-paced selection'' to train, starting with easy samples and progressively choosing harder samples. %\marcus{we should compare to them? if not why not?} 
Instead, we propose to create realistic data samples by segmenting, inpainting and blending the foreground of one video onto the background of another. %\ls{not sure if we should argue realism helps, let's see the ablation study}.
Crucially, we learn to discard novel video samples that are not expected to be useful for classification, overall producing a more accurate data augmentation strategy.

% VideoMix~\cite{videomix}
% ActorCut~\cite{actorcut}

%\ls{not sure if we need to compare to VideoMix in experiments?}

\paragraph{Learning to Augment Data.} 
The idea of learning to augment data has been used in other computer vision problems. In the image classification domain, this strategy has been done using the final classification loss as the training criterion~\cite{Lemley2017SmartAL}, augmenting in feature space~\cite{devries2017dataset}, and learning data augmentation policies~\cite{Cubuk_2019_CVPR}. As in this paper, in the image domain it has been noted that the search space for data samples can be  large and thus expensive~\cite{NEURIPS2020_auto}. 

Other computer vision domains like low level vision, also struggle with data dependency, as creating ground truth is particularly hard. In optical flow, AutoFlow~\cite{sun2021autoflow} recently introduced the strategy of learning to generate good training data for a target dataset.

\paragraph{Semi-supervised Video Action Recognition.} Semi-supervised learning (SSL) also aims to reduce data dependence by learning from large sets of unlabeled samples and a small set of labeled ones. SSL in images has been widely explored.
%\ls{are these all image techniques?}
For example, some strategies include giving pseudo-labels \cite{arazo2020pseudo,pseudo} to samples where the classifier has high confidence, and adding these to the labeled training data. Other common approaches use consistency regularization \cite{conreg1,conreg2,meanteacher}. %The motivation to use consistency regularization is because a model is expected to generate consistent predictions when using the same unlabeled data that has been subject to various transformations or perturbations. 
Approaches that combine consistency regularization and entropy minimization ~\cite{grandvalet2005semi} have shown to be very effective in tackling the SSL task in images such as MixMatch \cite{berthelot2019mixmatch} and RemixMatch \cite{berthelot2019remixmatch}.

SSL in videos however, has not been explored as much. One of the early works used extreme learning machines \cite{iosifidis2014semi} to perform SSL on videos. Recently, VideoSSL \cite{Jing_2021_WACV} and Temporal Contrastive Learning (TCL) \cite{tcl} leverage SSL in videos. %ActorCut \cite{actorcut} cuts out actors from one clip and paste it onto another clip and use multiple such clips to aid the training process. 
VideoSSL \cite{Jing_2021_WACV} uses pseudo-labels and object cues from unlabeled samples to guide the learning process. TCL \cite{tcl} use a two-pathway contrastive learning model using unlabeled videos at two different speeds with the intuition that changing video speeds do not change the action being performed.

Data augmentation and SSL are two different families of techniques to relieve the dependence on labeled data, and in this paper we experiment with the combination of both, showing that they are actually complementary. 

%Of these works, ActorCut~\cite{actorcut} is most similar to our work. However, they randomly pick foreground and background samples from an exponentially large search space and then paste the foreground actors on the background. We differ from them by making more informed choices of our foreground and background samples. Specifically, we train a selection network to decide what samples to pick and use semantically similar classes to drastically reduce our search space. Also, since actions performed are not just dependent on actors, but on certain objects, we remove objects along with actors from the foreground and background samples.

%\ls{is there any work on synthetic data?}

\paragraph{Sample Selection.}
Recent work~\cite{whatmakes} has shown that not all data samples are as useful. Selecting a subset of high quality frames or clips at test time shows better results than using the entire video for action recognition. In this spirit, SMART~\cite{smart} uses an attention and relation network to learn scores for each frame in a video and then select only the high ranked ones for inference. Similarly, SCSampler~\cite{korbar2019scsampler} uses a lightweight clip sampling model to select the salient clips in a video and use only those. Unlike the proposed method, these learn to choose single videos, which are already available, while we learn to choose pairs of videos to be composited, which are not already combined. 

The most relevant work to ours is data valuation in the image domain, using RL \cite{yoon2020data}, in the image domain where each sample is given a score of how effective the sample is, and at training time the sample is multiplied by this score. In our work, instead of learning the effectiveness of the training set, we leverage that knowledge for augmentation.

\section{\Method}

\begin{figure*}[t]
\centering
    \includegraphics[width=0.7\textwidth]{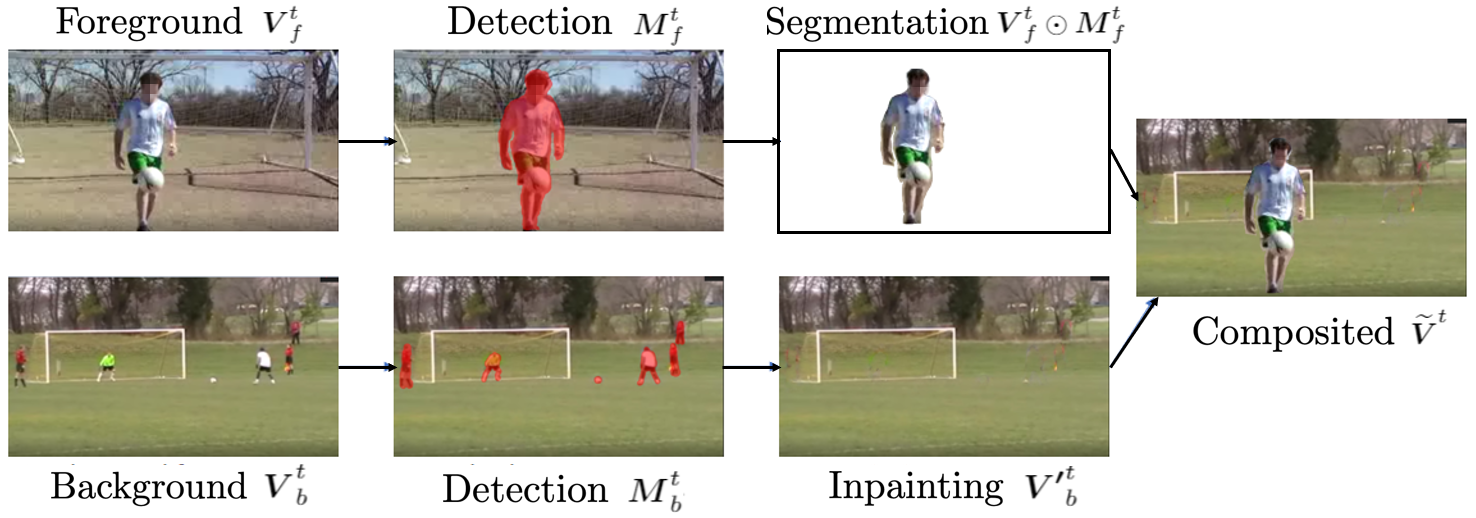}
    \caption{Pipeline for compositing a single frame. The foreground is from the class ``soccer juggling'' and the background from the class ``soccer penalty'', which are semantic class neighbors. We can see objects such as `person' and `ball' are detected as objects of interest.}
    \label{fig:mix}
\end{figure*}

%\marcus{I think this intro should focus more on the contribution, it might be good to narrow it down} 
%\ls{I'll come back to this, once the intro is done, and we know how to motivate. }

In this section we describe in detail the architecture of the proposed \Method. In a nutshell, the goal is to learn to augment novel data points which are realistic and diverse, such that we can train a better classifier with them. For this, we train a Selector network, which predicts a score of how useful a given pair of videos is for augmentation. We pick pairs that have a high score to be augmented. The transformation we use for augmentation is \VideoMix. Training the Selector using the entire dataset is infeasible, and sampling pairs of videos at random will yield unlikely pairs. Thus we sample pairs of videos using \SemanticMatch. Figure~\ref{fig:overview} shows an overview of the proposed method and in Sec.~\ref{sec:optimization} we describe how we train our approach.

% The proposed architecture for data augmentation has three components. First a Selector, which predicts how good the combination of two videos will be. Second a video compositing process, that combines two videos by segmenting the foreground of one and inpainting it on the foreground of the other. Third an action classification network. Figure~\ref{fig:overview} shows an overview of the proposed method. 

\subsection{Selector}
\label{subsec:choosing} 
% We make the hypothesis that a ``good'' training sample is one which, if used for training, improves the validation accuracy of the classification network. In other words, if we take one optimization step training the classification network, after updating the weights, the validation accuracy will go up. 

% In our context, this validation criterion would involve several steps in the na\''{i}ve approach: sample two videos, compositing them, input to the classification network, backpropagate and update the classification network and finally computing the validation accuracy. From all the steps, the compositing process is one of the most computationally expensive. In this section we describe how to predict whether a pair of videos will yield a good augmented video, without actually having to composite it. That way, we will only have to actually create the videos that will yield a ``good'' data sample. 

Given two input videos $\boldsymbol{V}_1$ and $\boldsymbol{V}_2$, the goal of the Selector is to predict a weight $\omega$, rating the quality of the potential composited video. Note that the input to the Selector is two putative videos instead of the composited one. This means that at test time, we can predict how useful the composited video will be without having to actually create it. 

The architecture of the Selector includes a standard video classification network to extract video features, which is ResNet3D-18~\cite{resnet} followed by a simple multi-layer perceptron (MLP) with 3 hidden layers of sizes 2048, 1024 and 512. Two videos are input to the Selector at a time, and their features and labels are concatenated and input to the MLP. 

Since there is no ground truth of how ``good'' a video sample is for learning, we train the Selector using the change in validation loss of the classifier. This is, we argue that a ``good'' training sample is one which, if used for training, improves the validation loss of the classification network. In other words, if we take one optimization step training the classifier, after updating the weights, the validation loss will go down. Section~\ref{subsec:RL} describes the training process in detail.

At test time, we use the Selector by sampling pairs of videos, choosing those pairs with high score $\omega$, and input to the \VideoMix module, which we describe in Sec.~\ref{subsec:video_mixing}. The resulting video is finally used to augment the training set for the classification network.

% The two videos are then input to the \VideoMix module, which we describe in Sec.~\ref{subsec:video_mixing}. The resulting augmented video is used to update the classification network with stochastic gradient descent. After the update, the network is evaluated on the validation set, to obtain a validation loss. If the new augmented video is ``good'', the validation loss will go down. We use this criterion to train the Selector network. We explain the optimization of the Selector in detail in Sec.~\ref{subsec:RL}.

%\myparagraph{Choosing Class Pairs.} 
\subsection{\SemanticMatch (SM)}

%\ls{Show samples of matched pairs somewhere.} 
%Note that although we can use the Selector network to choose pairs, 

The number of pairs in the full dataset can be very large, as it grows with the square of the number of videos. For Kinetics~\cite{carreira2017quo}, for example, we would encounter 360 billion pairs. Training the classifier using these is clearly infeasible, and thus we use the Selector. But training the Selector itself with all these samples is infeasible too. Sampling uniformly is a reasonable solution, but many video pairs may not be useful for learning. We leverage the observation that all combinations of actions and backgrounds are not equally likely~\cite{Choi-NeurIPS-2019}. This natural correlation between actions and backgrounds helps to prune unlikely class combinations. 

For this, we make the assumption that classes that are semantically similar are more likely to contain a foreground and a background that are plausible in the real world, and therefore more realistic for our data augmentation purposes. Thus, we use the class names to extract a language embedding using sen2vec \cite{sen2vec}, and use these embeddings to match each class to its nearest neighbor. We sample videos $\boldsymbol{V}_1$ and $\boldsymbol{V}_2$ from class $c_1$ and its closest neighbor $c_2$ respectively. This simple decision reduces the number of pairs to grow linearly with the size of the dataset, and furthermore increases the accuracy significantly with respect to sampling video pairs at random. More details on the numerical impact can be found in Sec.~\ref{subsec:ablation}. Semantic class pairs and additional experiments using intra-class augmentation can be found in the supplementary material.

\subsection{\VideoMix (VC)}
%\subsection{Segmented Object Mixing for Videos}
\label{subsec:video_mixing}

The goal of the augmentation process is to composite two videos, to produce realistic, plausible and diverse new videos, that will improve the classification. 
%We now describe the process of augmentation by compositing a pair of videos. 
Figure~\ref{fig:mix} shows the overall pipeline for compositing a single frame. 

Given two videos which will be used for foreground $\boldsymbol{V}_f$ and background $\boldsymbol{V}_b$, %\marcus{we should fix the nation here be use $\boldsymbol{V}_f$ above $v_1$. I like ''f'' and "b", but we should decide on captial vs not and make this consistent; also in the figures} 
we use a standard object segmentation network (MaskRCNN~\cite{maskrcnn}) to segment out people and objects in every frame of both videos. Objects categories in action datasets are not completely contained in the image dataset COCO~\cite{coco}, which is used for training MaskRCNN. However, we observe that object detections with high confidence tend to correspond to actual objects, even if the category is not correct (boxing bag is often classified as fire hydrant), and therefore are useful to our purpose. We could also have selected only the humans in the video, as action categories tend to be focused on humans. However, we find that the presence of specific objects is highly correlated with action categories (musical instruments in the classes ``playing guitar" or ``playing violin"). Therefore removing the original objects from the background and adding the ones from the foreground is essential for recognition. See numerical results of the impact of these decisions in the ablation study of Sec.~\ref{subsec:ablation}.

\begin{figure*}[t]
    \centering
    \includegraphics[width=0.13\textwidth]{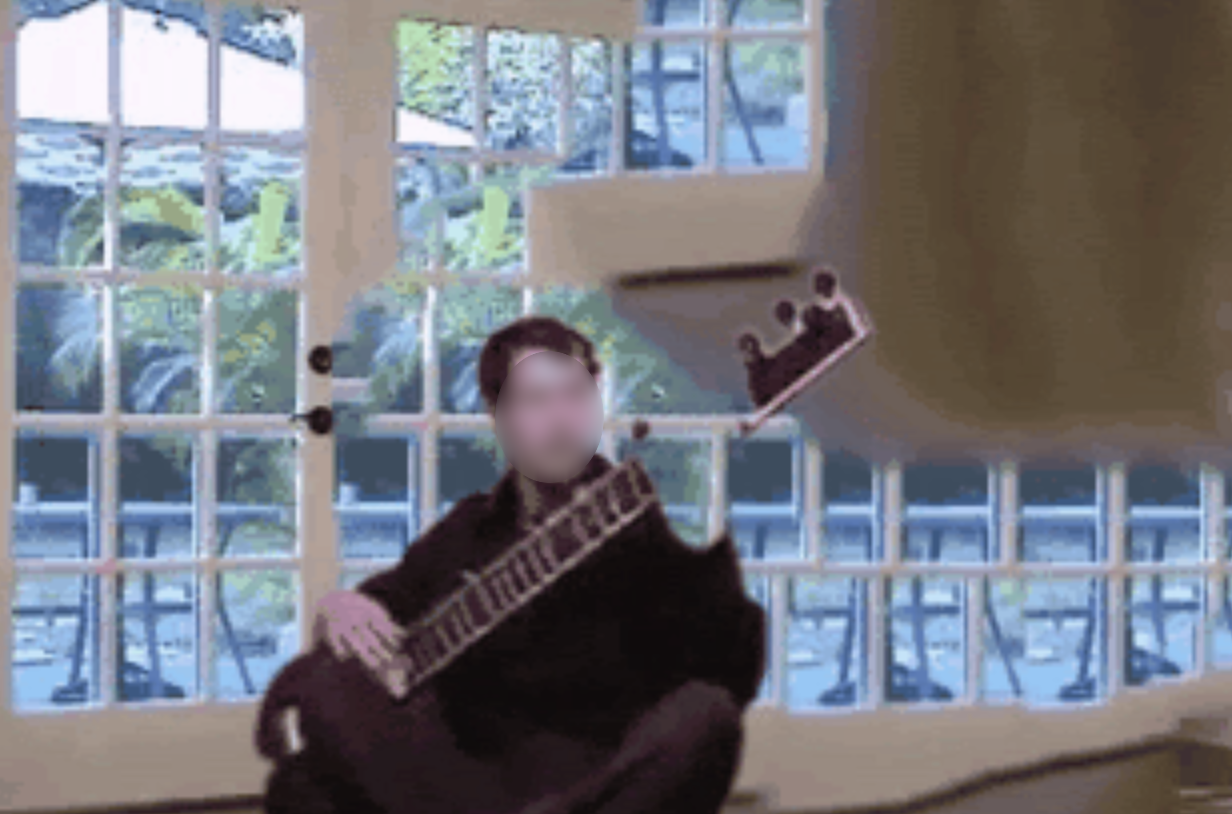}
    \includegraphics[width=0.13\textwidth]{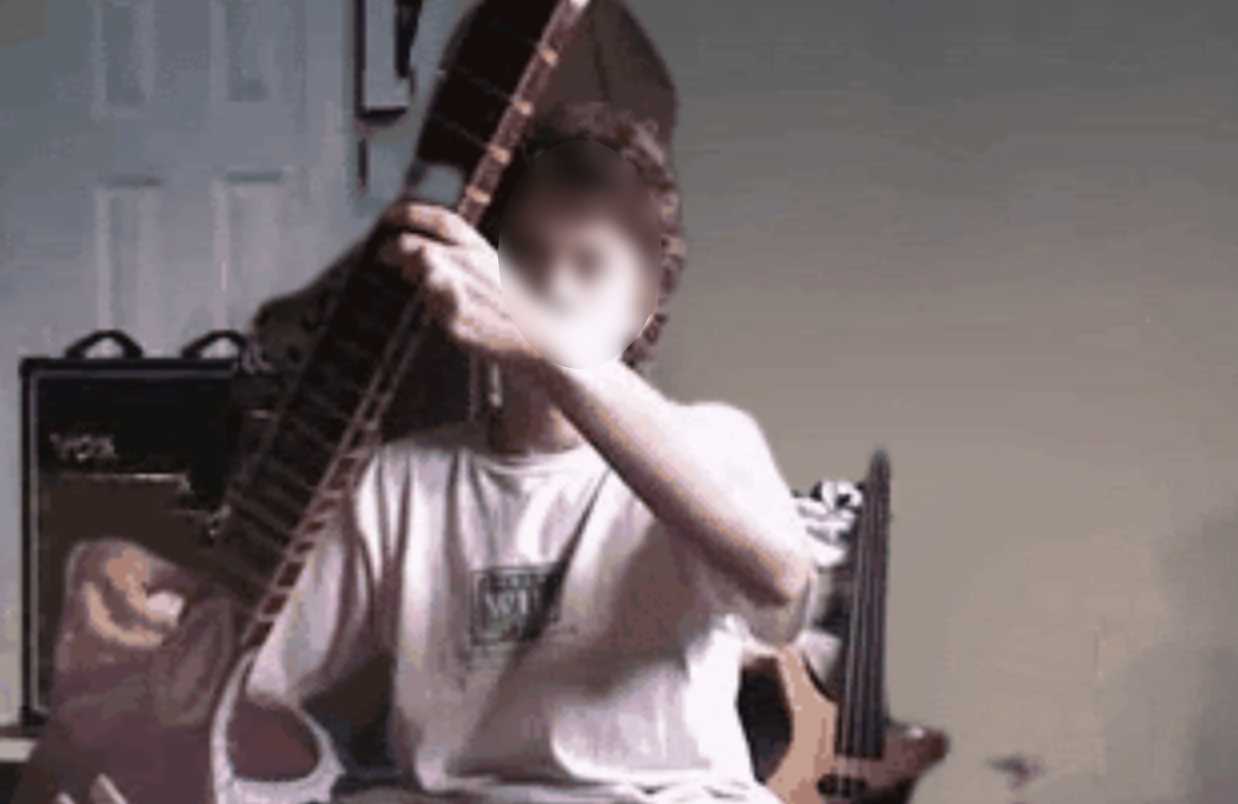}
    \includegraphics[width=0.13\textwidth]{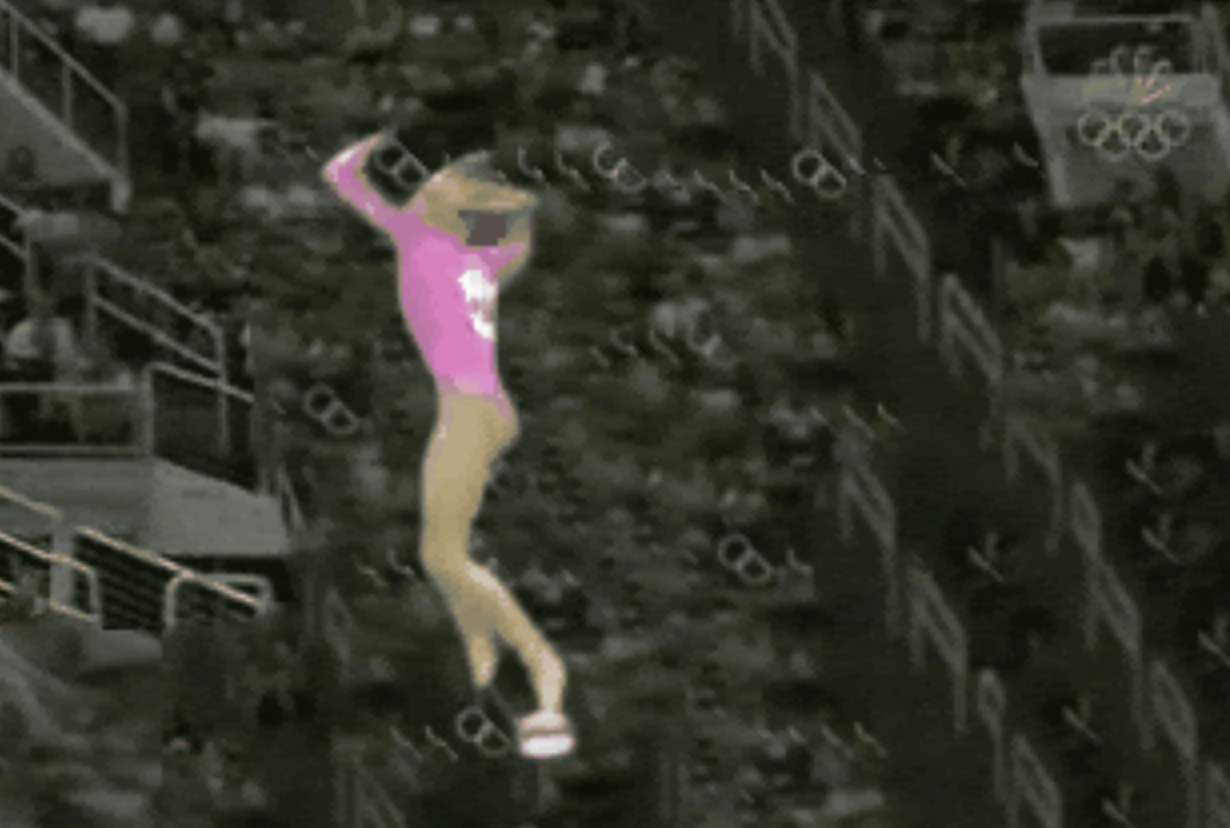}
    \includegraphics[width=0.13\textwidth]{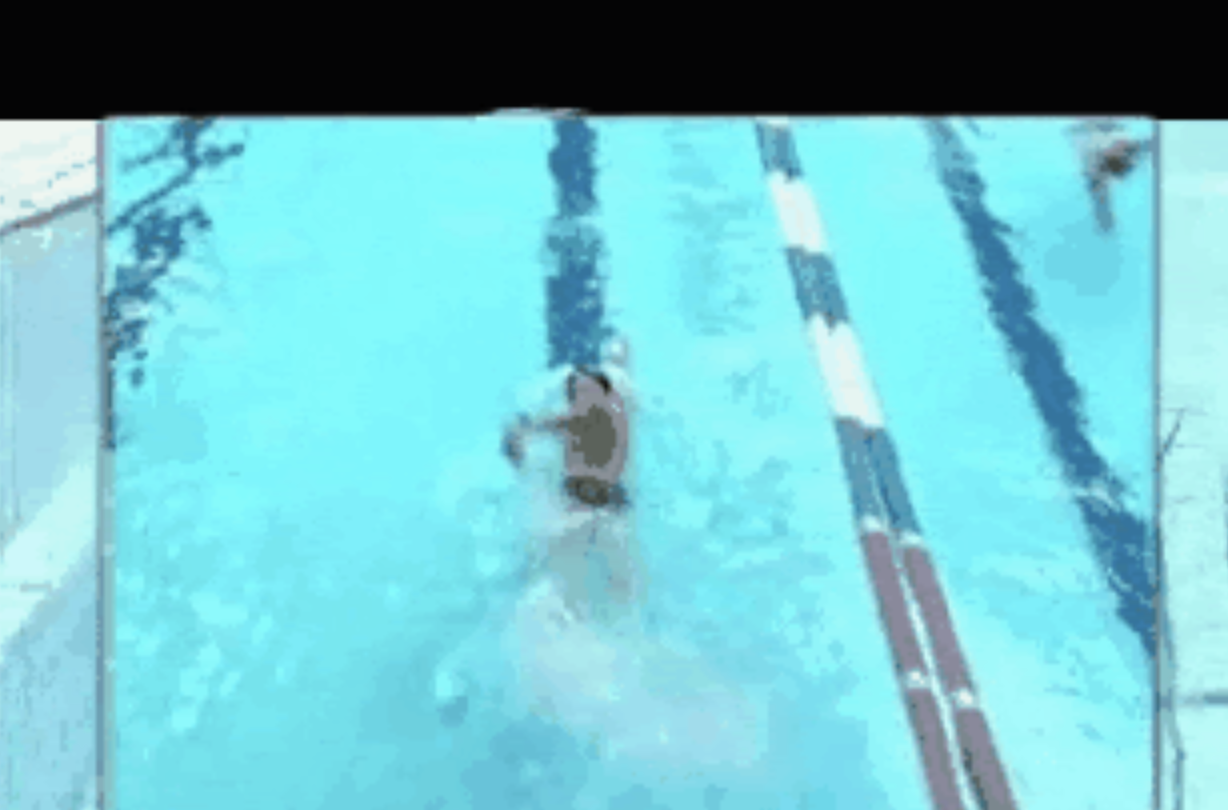}
    \includegraphics[width=0.13\textwidth]{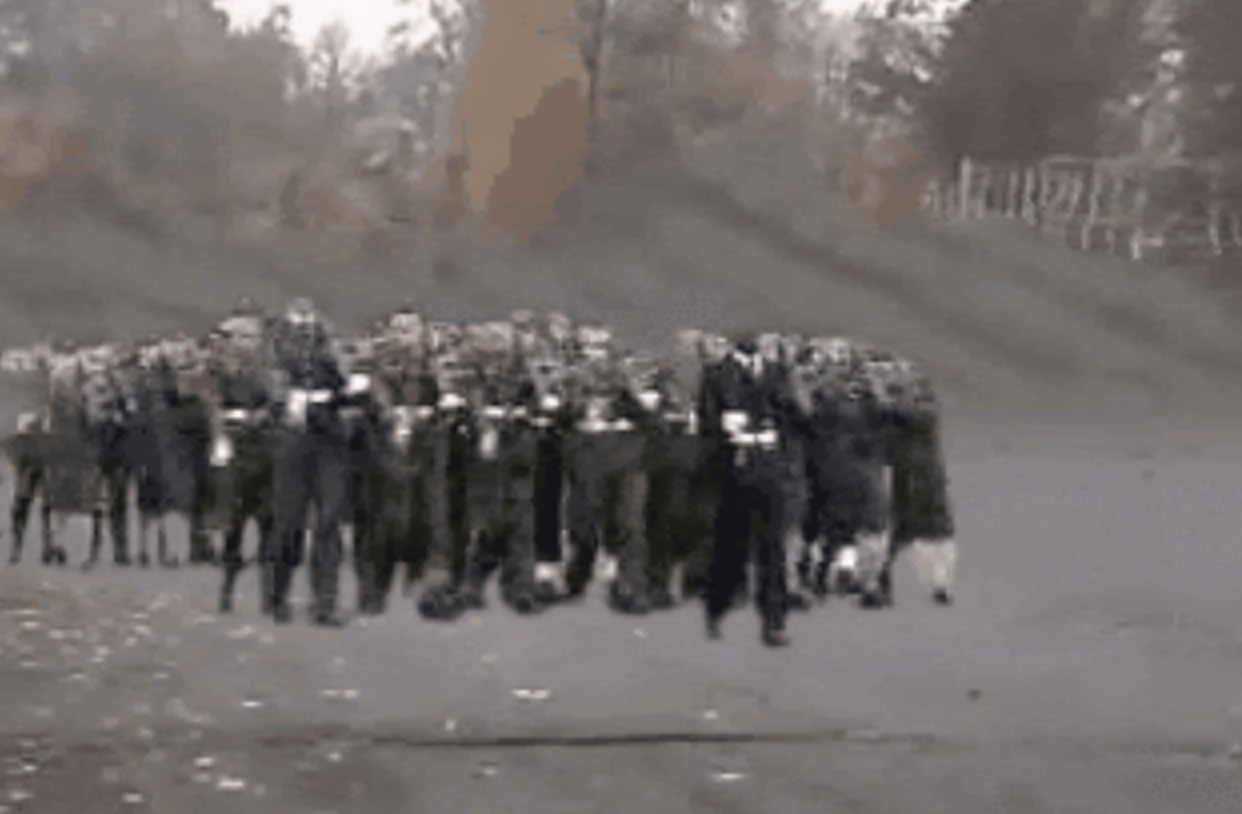}
    \includegraphics[width=0.13\textwidth]{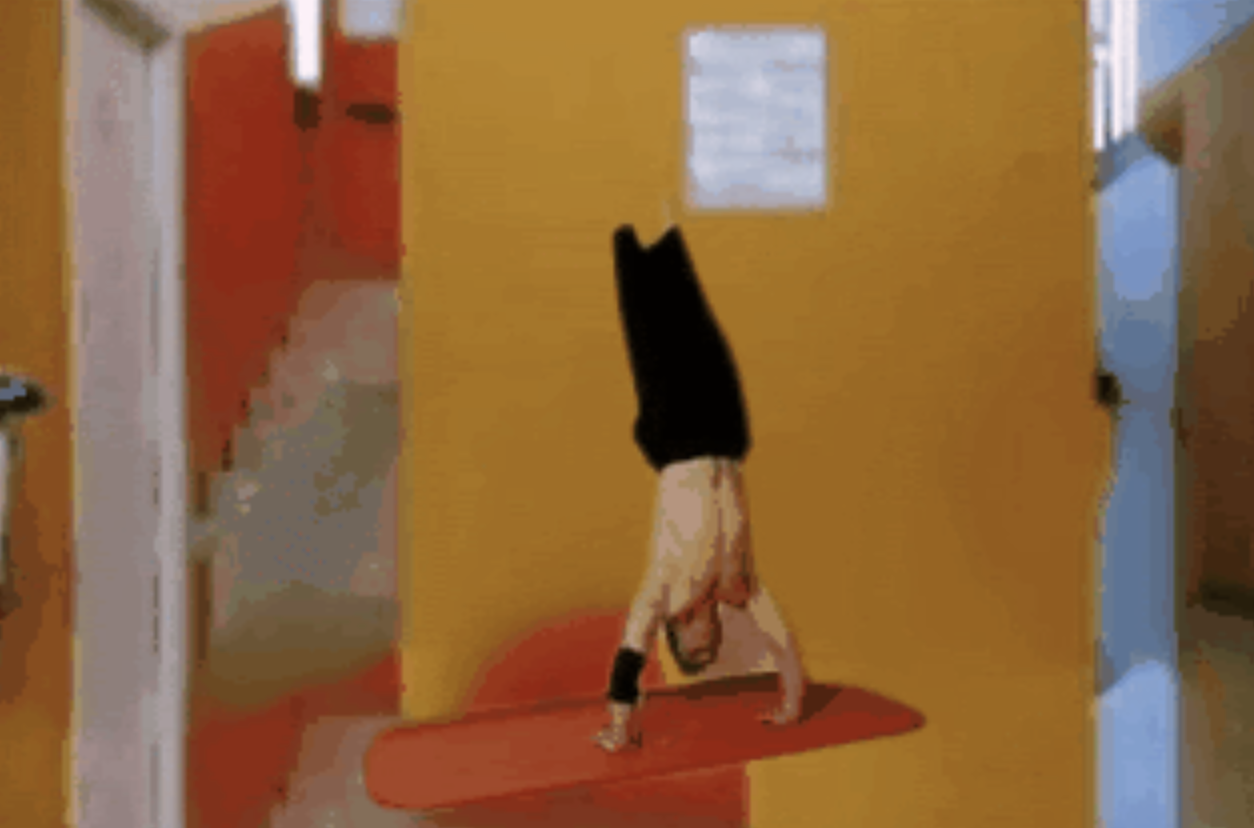}
    \includegraphics[width=0.13\textwidth]{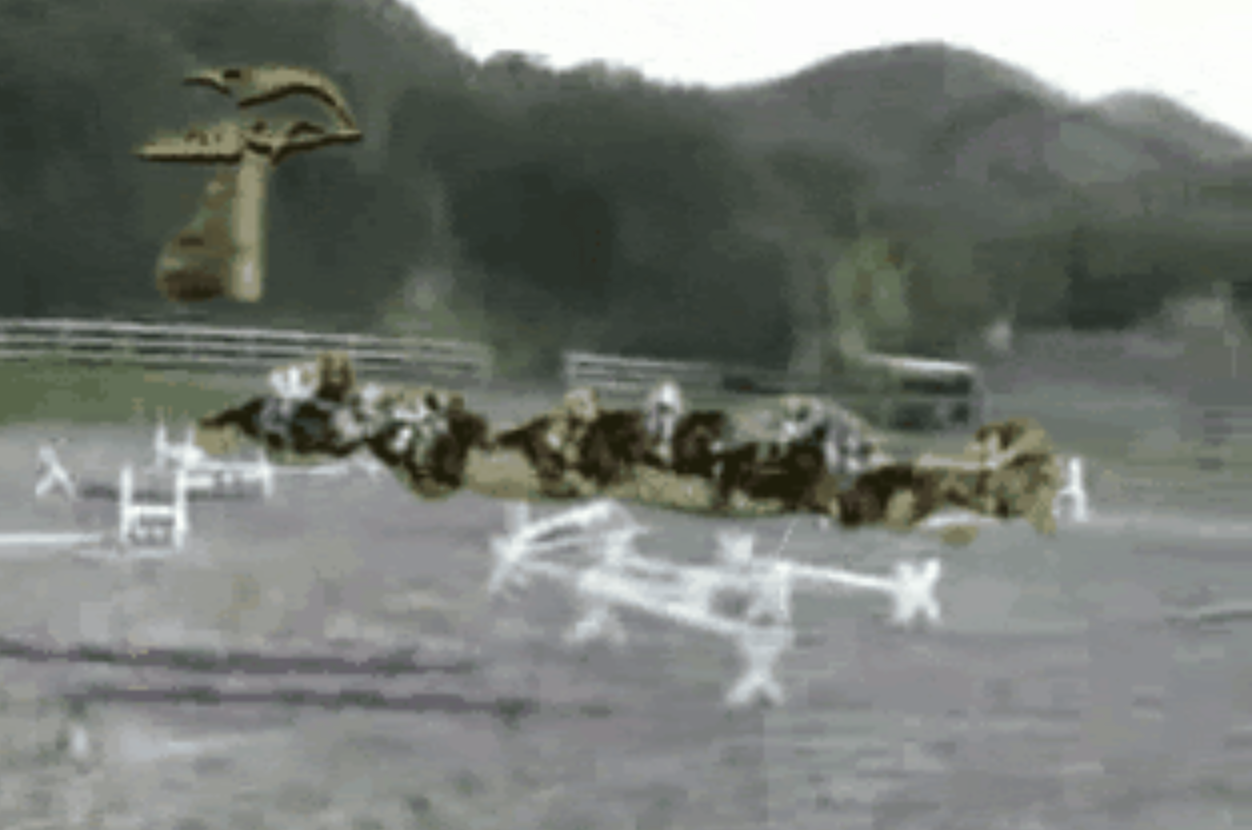}
    \caption{Sample frames of rendered videos. While the segmentation contains errors, such as missing limbs or portions of the object, the action category remains clear.}
    \label{fig:sample_videos}
\end{figure*}

We remove the segmented objects from the background video and fill in the holes using image inpainting~\cite{inpaint}, to obtain a clean background video $\boldsymbol{V'}_b$. Finally, we combine the foreground objects and the background at each frame by simple composition, as in: 
\begin{equation}
    \boldsymbol{\widetilde{V}}^t = \boldsymbol{V}_f^t \odot \boldsymbol{M}_f^t + \boldsymbol{V'}_b^t \odot (1-\boldsymbol{M}_f^t), 
\end{equation}
where $\boldsymbol{\widetilde{V}}^t$ %\marcus{add this symbol to the figure?} 
is the resulting composited frame at time $t$, $\boldsymbol{V}_f^t$ and $\boldsymbol{V'}_b^t$ are frames of the foreground and background videos respectively, $\boldsymbol{M}_f^t$ is the binary mask with the union of all detected objects, and $\odot$ is the element-wise multiplication. Figure~\ref{fig:sample_videos} shows sample frames of the resulting videos.

% \myparagraph{Object detection} with MaskRCNN~\cite{maskrcnn} 
% (person + what objects?) 

% \myparagraph{Inpainting}

% \myparagraph{Blending}

\section{Optimization of \Method}
\label{sec:optimization}
% \subsection{Optimization}

% \begin{enumerate}
%     \item Pre-train classification network
%     \item Train Selector network and classification network 
%     \item Train classification network 
% \end{enumerate}

The optimization of the proposed \Method method has two stages. In the first stage, we train the Selector network using RL, as described in Sec.~\ref{subsec:RL}. Once the Selector network is trained, in the second stage, we perform data augmentation to train the classifier. That is, we sample pairs of videos, pass them through the trained Selector, choose the pairs with high score, create new videos with these pairs through \VideoMix, and add them to the training set. We now describe the details of these two training stages.   

% In the first stage, we train the Selector network as we will describe in Sec.~\ref{subsec:RL}. At a high level, this training process involves mixing two videos, measuring how good they are for training, and using this criteria to update the Selector, until convergence. In the second stage, which we call the ``Data Generation Phase", we freeze the Selector network. We sample pairs of videos, input them to the Selector, and if their score is sufficiently high, we mix them and use the resulting augmented video sample to train the classification network. Once the classification network is trained, in the third and final stage, we test the classification network in the standard manner. 

\subsection{Training the Selector}
\label{subsec:RL}

As mentioned before, there is no ground truth to tell us how good an augmented data sample is. Instead, we use the validation loss of the classification network to train the Selector network. This function is not differentiable with respect to the parameters of the Selector. %validation accuracy calculated by a separate network makes optimization by gradient descent not feasible as the reward function is not differentiable with respect to the selection network. 
A common solution to dealing with this is to use RL \cite{yoon2020data}. 

Specifically, the state $s_t$ at time $t$ is the batch of video pairs sampled using \SemanticMatchShort. %\marcus{it might be good to say how you create the set of pairs.}. 
The action $a_t$ is the subset of these video pairs selected for compositing and is represented as a vector of values between 0 and 1.
The environment is the classification network and the validation process. This environment is used to compute a reward $R(\theta)$ for choosing a particular action, where $\theta$ are the parameters of the Selector. % \marcusAdd{$f_\theta$}. 
%, which is subset of pairs of videos, given the state, which is the entire batch of videos $s_t$ at time $t$. %This  (action)\marcus{maybe introduce here already the variables?}, .

We calculate the reward in a single step, as the difference between the loss in the current batch and the moving average of losses in the previous $S$ steps (where $S = 5$) denoted as $\delta$, as in Eq.~\ref{eq:reward}: 
\begin{equation}
\centering
R(\phi) = \left (  \frac{1}{|D_{\text{val}}|}\sum_{ i = 1}^{|D_{\text{val}}|}\mathcal{L}_\text{cls}(f_{\phi }(\boldsymbol{V}_{i}),y_{i})\right ) - \delta 
\label{eq:reward}
\end{equation}
where $\mathcal{L}_\text{cls}$ is the classification cross-entropy loss, $f_{\phi}$ is the classifier network of parameters $\phi$, $\boldsymbol{V}_i$ and $y_i$ are an input video and its label respectively, $D_\text{val}$ is the validation set and $|D_{\text{val}}|$ is the number of samples in $D_\text{val}$. The objective function that we want to maximize is the expected value of the reward: 
\begin{equation}
\centering
%J(\theta) = \mathbb{E}_{i \in D_\text{val}}(\mathcal{L}_\text{cls}(f_{\phi }(x_{i}),y_{i}) - \delta)
J(\theta) = \mathbb{E}(R(\phi)).
\label{eq:objective}
\end{equation}
%
%where $\theta$ are the parameters of the Selector. 
To find the optimal policy, we would typically differentiate the objective function with respect to the parameters $\theta$. However, the reward function is dependent on the validation loss, calculated with the classifier network, which does not involve $\theta$. Instead, using REINFORCE~\cite{reinforce}, we approximate the objective function as: 
\begin{equation}
\centering
\nabla _{\theta }J(\theta)\approx \frac{1}{M}\sum_{i=1}^{M}R_{\tau^{i}}(\phi)\left( \sum_{t=0}^{T-1}\nabla _{\theta }\log \pi_{\theta}(a_{i}^{t}|s_{i}^{t}) \right),
\label{eq:approx}
\end{equation}

where, $\tau^{i}$ is the $i^{\text{th}}$ state-action trajectory under the policy $\pi_{\theta}$, $M$ is the number of sample trajectories and $T$ is the number of actions performed in a trajectory. Note that as we have single-step episodes, we can make several simplifications as $M=1$, as $T=1$, and as there is only one trajectory $\tau^{i}$, and thus $R_{\tau^{i}}(\phi)$ is just $R(\phi)$.
% $M$ is 1 in our c as we have single-step episodes) and this ensures only single-step rewards and hence $R_{\tau^{i}}(\phi)$ is just $R(\phi)$ for us, %\marcus{is k=1 for us, what is it for us?}, 
% and $T$ is the number of actions performed in a path, which again is 1 in our case as we have single-step episodes. %\marcus{I think for this to make sense not just be copy from the textbook we have to say what these things are in our case exactly}
%
%A state in our scenario is a batch of examples consisting of pairs of videos. The action is then choosing which of these pairs to mix. The policy returns the value of how good an augmented sample is. 
%During each episode a random batch is sampled, to obtain a reward on how good this batch is for classification. Since this happens for only one timestep, there is no transition function. Using this, we can approximate Eq~\ref{eq:objective} to Eq~\ref{eq:rl_final} using Eq~\ref{eq:approx}. 
%
With these simplifications and substituting Eq.~\ref{eq:objective} in  Eq.~\ref{eq:approx}, we obtain: 
\begin{equation}
%\begin{multline}
\centering
%\nabla _{\theta }J(\theta)\approx \left (  \left [  \frac{1}{K}\sum_{ \forall i \epsilon D_\text{val}}^{}\mathcal{L}_\text{cls}(f_{\phi }(x_{i}),y_{i})\right ] - \delta \right ) \nabla _{\theta }\log\pi(D_{M}|D_{B}), 
\nabla _{\theta }J(\theta)\approx R(\phi)\nabla _{\theta }\log\pi_{\theta}(D_{M}|D_{B}),
\label{eq:rl_final}
%\end{multline}
\end{equation}
where $D_{M}$ corresponds to the subset of pairs of samples to composite and $D_{B}$ to all the pairs of samples in the batch. %Hence, $D^{M}\subseteq D^{B}$.
The Selector is updated by $\alpha\nabla _{\theta }J(\theta)$ where $\alpha$ is the learning rate and $\delta$ is updated with the last calculated loss as seen in Eq~\ref{delta}.
%We use the REINFORCE algorithm \cite{reinforce} to optimize using policy gradients. %The rewards are obtained from a validation set. 

\begin{equation}
    \delta_t = \frac{S-1}{S}\delta_{t-1} + \frac{1}{|D_{\text{val}}|}\sum_{i=1}^{|D_{\text{val}}|} \left ( \mathcal{L}_{cls}(f_{\theta}(\boldsymbol{V}_{i}),y_{i}) \right ).
\label{delta}
\end{equation}

Note that this training process does involve generating the composited videos for pairs in $D_M$, to input to the classifier and compute the loss. However, crucially, during training this is a small portion (one order of magnitude smaller) of how many videos would need to be generated if we were to composite all pairs of videos. %In addition, not only choosing the videos to combine is much faster, it also leads to better accuracy, as the chosen videos are better for training. Section~\ref{sec:experiments} shows more concrete numerical results. 

%\marcus{One could add a connection here how the semantic matching reduces the action space meaningfully and thus makes the learning easier.}

Once the Selector is trained, we use it for actually filtering good pairs. At that point, given two videos and their labels, the Selector network predicts a policy $\pi$ of how likely it is to select the pair. The score $\omega$ is the value of $\pi$ for each pair. We use a threshold on that score to select the pairs of videos to augment. In our experiments, we first determine a budget on the number of videos that we want to augment, and then pick the threshold to select the top-ranked video pairs. We use these selected pairs of videos as input to \VideoMix, add them to the training set, and use them to train the classifier. 

\subsection{Training the Classifier}

%\myparagraph{Label Smoothing.}

Similar to previous work which combines multiple samples for augmentation~\cite{yun2019cutmix,actorcut}, composited/mixed samples should include mixed labels. We adopt the strategy of Cutmix~\cite{yun2019cutmix}, where the foreground label $y_f$ and the background label $y_b$ are combined using a ratio $\lambda$, as:
\begin{equation}
    \tilde{y} = \lambda y_{f} + (1-\lambda)y_{b},
\label{eq:cutmix}
\end{equation}
to obtain the mixed label $\tilde{y}$. A simple way to choose $\lambda$ is to use the ratio of the foreground mask with respect to the overall video. Given the foreground video $\boldsymbol{V}_f$ of dimensions $T \times H \times W$, and mask at each frame ${M}_{f}$, the foreground ratio would be $\gamma = \sum \boldsymbol{M}_{f}/({THW})$.
% $\boldsymbol{V}_f\in R^{T \times H \times W \times 3}$, we can compute this as:
% \begin{equation}
%     \gamma = \frac{\sum \boldsymbol{M}_{f}}{THW},
% \label{eq:fg}
% \end{equation}
% where $\boldsymbol{M}$ is the foreground mask. 
%Label smoothing is used in CutMix to provide a weighted representation of multiple labels in a single training image as that image contains information of two different classes. 
Instead of choosing $\lambda$ to be directly proportional to the foreground ratio $\gamma$, we give slightly more weight to the foreground~\cite{actorcut}, as in Eq.~\ref{eq:scaling}, where $\alpha = 4$.
%
%. We choose $\alpha = 4$. \marcus{is 4 different from prior work?}
%
\begin{equation}
    \lambda = -(\gamma -1)^{\alpha} + 1, \gamma \in [0,1]
\label{eq:scaling}
\end{equation}
We add composited videos $\boldsymbol{\widetilde{V}}$, and their mixed labels $\tilde{y}$ to the training set, and train the classifier network using a standard cross-entropy loss, with stochastic gradient descent. 

The choice of classifier is not tied to our method. In our experiments, we choose the widely used 3D ResNet-18 architecture, which allows us to compare directly to other approaches.

% However, actions tend to be more weighted by the foreground rather than the background. Hence, we follow \cite{actorcut} to give a lower weight for the background sample with the help of a scaling function as seen in \ref{eq:scaling}. 

\section{Experiments}
\label{sec:experiments}

We experiment extensively with \Method using three data settings, four datasets, and two splits. We also present ablation studies. In this section we first describe the details of the experiments and then discuss our results.

\subsection{Experimental Details}

\paragraph{Datasets.} In order to provide comparison to prior work (e.g. \cite{actorcut,tcl}), we use standard datasets for evaluation in action recognition, including HMDB51~\cite{hmdb}, UCF101~\cite{ucf}, Kinetics-400~\cite{carreira2017quo}, and Kinetics-100, which includes the 100 classes with the largest amount of samples in Kinetics, as it is used in prior work~\cite{Jing_2021_WACV} and helps us compare directly. For experiments on the effect of pre-training the Selector, we use Kinetics-400. For the semi-supervised setting, we split the datasets following the protocol of  VideoSSL~\cite{Jing_2021_WACV} and ActorCut~\cite{actorcut}. %UCF101 has 101 classes with 13,320 videos, HMDB-51 consists 51 classes with 6766 videos. Kinetics-400 consists of 400 classes and 300K videos, while the subset used Kinetics-100 has 100 classes and 90K videos.
For few-shot we use the standard split \cite{arn} and the Truze split \cite{truze} which ensures no overlap of novel classes with Kinetics-400.

\paragraph{Problem Settings.} We test the proposed method in three different settings. In the \emph{semi-supervised} setting, a portion of the training set is artificially held out, and the rest of the training data is assumed to be available, but unlabeled. Tests are performed on different percentages of held out data. In the \emph{few-shot} setting, some classes (novel classes) are assumed to have a very small number of training samples (one to five instances), while other classes have the full number of samples (seen classes). We effectively change the $n$-shot learning problem to a $n+k$-shot problem where $k$ is the number of augmented samples. 
%In this setting, we have a large number of labeled examples for some classes (seen classes) and only a few examples for other classes (novel classes). 
Finally, in the standard \emph{full set} setting, all training data is available. 
% \myparagraph{Architecture.} 

% The Selector network is an MLP with 3 hidden layers of sizes 2048, 1024 and 512. The input to the Selector network is concatenated feature representations of two random samples $x_{i},x_{j}$ and the output is a score telling how good the mixed sample is. We implement our method on
% top of the publicly available mmaction2 codebase\footnote{https://github.com/open-mmlab/mmaction2} 

\paragraph{Training Settings.} We use mini-batch stochastic gradient descent, with momentum of 0.9 and weight decay 0.001. For each video, we use an 8-frame clip, where the frames are uniformly sampled. We use batch size of 8. For UCF101 and Kinetics100 in the SSL setting, we train the model for 400 epochs and for HMDB51, we train for 500 epochs. The initial learning rate is set to 0.1 and then decayed using cosine annealing policy. For the SSL setting, we use the data split proposed in VideoSSL \cite{Jing_2021_WACV}. % where a random set of samples are chosen to have varying percentages of labeled data. 
For the few-shot setting, we use the default hyperparameters of TRX~\cite{trx}, ARN~\cite{arn} and C3D-PN~\cite{pn}, respectively. In the fully supervised setting, we train R(2+1)D for 100 epochs on UCF101, HMDB51 and 50 epochs on Kinetics-400.

%We train the classification network using a standard cross-entropy loss. In the experiments which involve using unlabeled samples, we adopt the VideoSSL~\cite{Jing_2021_WACV} architecture. 

%As in previous work that combines multiple data samples to create new ones, we use label smoothing. In particular, we adopt the same strategy as ActorCut~\cite{actorcut}, where the two labels that contribute to the final ground truth are weighted by the ratio of pixels that each take up, but giving a larger weight to the foreground. \ls{not sure if we want to give more detail here to make it self-contained or we simply refer to the other paper...}

% \begin{equation}
%     \lambda = 
% \end{equation}

%For training~\ls{check this} we use a single 8-frame clip per video, where these 8 frames have been sampled at random from the original video. All training is done with batch size of 8. The learning rate is~\ls{to fill in}. 

%The hidden layers of the MLP are of sizes 2048, 1024 and 512. 

\subsection{Architectural Changes for Different Settings}

%We show experimental results of the proposed augmentation method on three different settings: SSL, few-shot learning (i.e., 1-5 training samples per class) and the standard learning with full dataset. 
We briefly explain the structural adaptations of our approach for each of the settings.

\paragraph{Semi-supervised Learning.}
%We make use of both labeled and unlabeled data during training 
Similar to VideoSSL \cite{Jing_2021_WACV}, we first train the classifier on the available labeled data using the categorical cross-entropy loss. Once this network is trained, we do a forward pass of the unlabeled examples and assign pseudo-labels to those samples with high confidence. We use these pseudo-labels as additional data for augmentation. We also add a knowledge distillation loss inspired by VideoSSL \cite{Jing_2021_WACV}. Details can be found in the supplementary material.

\paragraph{Few-shot Learning.}
We only augment the novel classes using \Method. We also do not perform label mixing and simply use the foreground label for the augmented sample. This incorporates our composited samples seamlessly into the meta-learning framework typically followed. 
%Depending on the number of augmented examples, we essentially convert a n-shot problem into n+k-shot problem where `k' is the number of augmented samples that we add. We use our Selector network pre-trained on the entire Kinetics-400 dataset to help pick pairs of samples for mixing. 
We show results on the standard split, as on the recently proposed TruZe \cite{truze}. TruZe ensures that the novel classes do not overlap with Kinetics-400.

\paragraph{Fully-supervised Learning.} This is the simplest setting, where the Selector is trained on the full training set, and used for data augmentation to train the classifier. We explore two scenarios: training the classifier from scratch and using a model pre-trained on Sports1M \cite{sports1M}.

%\myparagraph{Label Smoothing}
%\marcus{this is a known technique and we don't innovate here, let's move it to implementation details?}

%\myparagraph{Incorporating Unlabeled Data. }
%\marcus{this is the setup from prior and we don't innovate here, let's move it to experimental setup/the experiment section which is about this setup?}

\setlength{\tabcolsep}{4pt}
\begin{table}[t]
\centering
\begin{tabular}{c c c c c}
%\hline %\noalign{\smallskip}
Pairs  & Video  & Semantic  & Accuracy & \#Videos  \\
 Selector &  Compositing &  Matching & in \% &  (S) \\
\noalign{\smallskip}
\hline
%Traditional Augmentation & 33.0 & 10442K \\ %These are possible samples to look at and not actual samples looked at as it's randomly chosen in traditional aug, plus no pseudo labels used in this case
%ActorCut & 45.5 & 10442K \\
%ActorObjectCut & & \\
%Mixing with Bounding Box & & \\
%ActorObjectCut + VideoMix & & \\
%Ours + Semantic Match (SM) &  55.8 & 99K \\
%Ours + SM + Label Smoothing (LS) & 72.5 & \\
%Ours with label smoothing (Discriminator selection) & 70.9 \\
%Ours + SM + LS + SMART~\cite{smart} & 72.8 & \\
%Ours + SM + RL selection & \textbf{58.9} & 12K  \\\hline
\checkmark & \checkmark & \checkmark & {\bf 58.9} & 12K  \\
$\times$ & \checkmark & \checkmark & 55.8 & 99K  \\
\checkmark & $\times$  & \checkmark & 54.5 & 12K  \\
\checkmark & \checkmark & $\times$  & 55.2 & (1.2M) \\
\checkmark & $\times$ & $\times$ & 52.9 & (1.2M) \\
$\times$ & \checkmark & $\times$ & 48.6 & (10.4M) \\
$\times$ & $\times$ & \checkmark & 50.8 & 99K \\
$\times$  & $\times$  & $\times$  & 45.5 & (10.4M) \\ % similar to actor cut and no selection
%\specialcell{Traditional} & 
\noalign{\smallskip}
\end{tabular}
\caption{Ablation study to explore the impact of each proposed component. All settings use the same number of samples for training, so that they can be compared fairly. The \# Videos (S) corresponds to the search space in each scenario. As we can see, we obtain the best accuracy using just 12K instead of the standard scenario which would have had 10.4M i.e. a reduction of over 1000x. }

\label{table:ablation}
\end{table}

\subsection{Ablation Study}
\label{subsec:ablation}

Table~\ref{table:ablation} shows the ablation study of \Method, which illustrates the impact of each of the proposed elements in the design. The experiment is done on the UCF101 dataset, using 20\% of the data i.e. in a semi-supervised setting. All three contributions (Selector, \SemanticMatch and \VideoMix) improve accuracy. Crucially, \SemanticMatch and the Selector also reduce greatly the number of possible video combinations, and the overall reduction is around three orders of magnitude. We see that \Method obtains a 13.4\% improvement over the baseline. While there are improvements of up to 7.4\% for each component, the combination of all three gives the best results. Further analysis can be found in the supplementary material.

The \VideoMix module also has multiple components. In Table~\ref{table:ablation_mix}, we ablate these components and observe that removing objects is actually essential, and has the most significant impact, followed by using segmentation instead of a bounding box, and finally inpainting. 

Although the compositing process is more computationally expensive than previous simpler mixing strategies, it is important to note that 1) the overall accuracy indeed improves, 2) the actual composition for training the classifier is done on a small subset of pairs of videos and 3) the Selector can be trained on a large dataset (e.g.: Kinetics) just once and can be reused for the smaller datasets without the need of fine-tuning (see Table~\ref{tab:semi}). 

\begin{table}[t]
\centering
\begin{tabular}{lc}
Method                  & Accuracy \\
\hline
\MethodShort                    & 58.9     \\
\MethodShort w/o Inpaint        & 57.6     \\
\MethodShort w/o Segmentation       & 56.8     \\
\MethodShort w/o Objects & 55.7    \\
\MethodShort w/o All & 54.5 \\
\end{tabular}
\caption{Ablation study of compositing components. The version ``w/o Inpaint" refers to pasting the foreground without first filling in the holes of removed objects in the background. The version ``w/o Segmentation" refers to using bounding boxes instead of object segmentations. ``w/o Objects" refers to copying and pasting only the humans in the scene, leaving the objects. }
\vspace{-0.5cm}
\label{table:ablation_mix}
\end{table}

%\subsection{Augmenting Limited Training Data}

\begin{table}[]
\setlength{\tabcolsep}{5pt}
\centering
\resizebox{\columnwidth}{!}
{\begin{tabular}{l@{\ }cccccccccccc}
     &  & \multicolumn{4}{c}{Kinetics 100} & \multicolumn{4}{c}{UCF101} & \multicolumn{3}{c}{HMDB51} \\
Method & Conference & 50\%   & 20\%   & 10\%   & 5\%   & 50\%  & 20\%  & 10\% & 5\% & 60\%    & 50\%    & 40\%   \\
\hline
CutMix~\cite{yun2019cutmix} & ICCV19 & 53.7 & 46.1 & 43.2 & 39.9 & 46.1 & 36.5 & 34.6 & 25.8 & 33.9 & 30.8 & 27.8 \\
MixUp~\cite{mixup} & ICLR18 & 53.4 & 45.5 & 43.0 & 39.6 & 45.8 & 36.1 & 34.2 & 25.5 & 33.7 & 31.0 & 27.5 \\
CutOut~\cite{cutout}& Arxiv17 & 52.8 & 45.1 & 42.3 & 38.8 & 45.2 & 35.6 & 33.9 & 24.6 & 33.0 & 30.5 & 27.1 \\
ST-VideoMix~\cite{videomix} &Arxiv21 & 55.3 & 46.6 & 43.9 & 40.4 & 46.4 & 36.4 & 35.2 & 25.9 & 34.8 & 31.3 & 28.7 \\
\hline
PseudoLabel~\cite{pseudo} &ICMLW13  & 59.0 & 48.0 & 38.9 & 27.9 & 47.5  & 37.0  & 24.7 & 17.6 & 33.5    & 32.4    & 27.3    \\  
MeanTeacher~\cite{meanteacher}& Neurips17 & 59.3 & 47.1 & 36.4 & 27.8  & 45.8  & 36.3  & 25.6 & 17.5 & 32.2    & 30.4    & 27.2    \\
S4L~\cite{s4l}& ICCV19 & 54.6 & 51.1 & 43.3 & 33.0 & 47.9  & 37.7  & 29.1 & 22.7 & 35.6    & 31.0    & 29.8    \\
VideoSSL~\cite{Jing_2021_WACV}& WACV21 & 65.0 & 57.7 & 52.6 & 47.6  & 54.3  & 48.7  & 42.0 & 32.4 & 37.0    & 36.2    & 32.7    \\
ActorCut~\cite{actorcut} &Arxiv21 & 68.7 & 61.2 & 56.8 & 52.7  & 59.9  & 51.7  & 40.2 & 27.0 & 38.9    & 38.2    & 32.9    \\
ActorCut+ID~\cite{actorcut}& Arxiv21 & 72.2 & 68.7 & 63.9 & 59.1 & 64.7 & 57.4 & 53.0 & 45.1 & 40.8 & 39.5 & 35.7    \\
TCL~\cite{tcl}& ICCV21 & 70.4 & 64.7 & 61.1 & 58.2 & 62.1 & 55.4 & 52.1 & 42.8 & 41.2 & 40.4 & 34.8\\
\MethodShort& & \textbf{75.9} & \textbf{72.1} & \textbf{67.5} & \textbf{63.7} & 72.1 & 60.3 & 56.1 & 48.0 & 44.5 & 43.2 & 37.9 \\ 
\hline
\MethodShort+Pre-training& & - & - & - & - & \textbf{73.3} & \textbf{64.8} & \textbf{60.1} & \textbf{50.9} & \textbf{47.1} & \textbf{46.3} & \textbf{42.1} \\
\end{tabular}}
%\marcus{* for TCL not explained}
\caption{Results on the semi-supervised setting. Results for TCL and ActorCut are obtained by us running the author's code. All methods are run with a 3D ResNet-18 backbone for fair comparison. \MethodShort+Pre-training refers to pre-training the selector and fixing it.}

\label{tab:semi}
\end{table}

\subsection{Augmenting in the Semi-supervised Setting}

In this setting we artificially hold out a portion of the training set, with the goal of observing the behavior of different methods as the size of the training set changes. In this setting, we use the remaining part of the dataset by producing pseudo-labels, similar to VideoSSL~\cite{Jing_2021_WACV}. Table~\ref{tab:semi} shows results in this semi-supervised setting. The \MethodShort version of the method uses a Selector and a classifier trained only on the target dataset (in this case UCF101, HMDB51 or Kinetics-100). We observe that \Method improves on all settings over all previous methods. 

The ``\MethodShort+Pre-training'' row refers to \Method where the Selector has been pre-trained on Kinetics-400, without fine-tuning on the target dataset. 
%\marcus{Does prior work use pre-trained representation on kinetics? we should note that if the representation is pre-trained or not}
We make two observations: First that pre-training on a large dataset helps, as the results from the pre-trained model are higher for all datasets and settings. Second that the Selector trained on Kinetics generalizes quite well to the smaller datasets without the need for fine-tuning. We do not test on Kinetics-100 with the pre-trained model, as this would mix training and testing sets.

\subsection{Augmenting in the Few-shot Setting}

We also explore the impact of the proposed method on the more extreme few-shot setting, where there are only a few examples per class. This is interesting because few-shot methods are already designed to address data scarcity. 

We compare with the current state of the art in this setting, including CD3-PN~\cite{pn}, ARN~\cite{arn} and TRX~\cite{trx}, on the UCF101 and HMDB51 datasets. %We use the standard split and the recent TruZe~\cite{truze} split, which removes the overlap between pre-training classes and test classes.
We observe that the proposed \Method method improves upon all existing approaches, suggesting data augmentation is complementary to few-shot methods. Table~\ref{tab:fslucf} shows the results of the experiments. 
 
\begin{table}
\centering
\resizebox{\columnwidth}{!}
{
\begin{tabular}{l@{\ }ccccccccccc}
     &  & \multicolumn{5}{c}{UCF101} & \multicolumn{5}{c}{HMDB51}\\
Method & Split & 1 & 2 & 3 & 4 & 5 & 1 & 2 & 3 & 4 & 5\\
\hline
%C3D-PN \cite{pn} & 78.2 & 75.4 & 2.8 & 57.4 & 49.1 & 8.3\\
C3D-PN \cite{pn} & S & 57.1 & 66.4 & 71.7 & 75.5 & 78.2 & 38.1 & 47.5 & 50.3 & 55.6 & 57.4 \\
C3D-PN + \MethodShort & S & \textbf{60.8} & \textbf{68.9} & \textbf{73.3} & \textbf{76.6} & \textbf{79.1}& \textbf{39.8} & \textbf{48.9} & \textbf{51.5} & \textbf{57.3} & \textbf{58.2} \\
\hline
%ARN \cite{arn} & 83.1 & 80.5 & 2.6 & 60.6 & 53.2 & 7.4\\
ARN \cite{arn} & S & 66.3 & 73.1 & 77.9 & 80.4 & 83.1 & 45.5 & 50.1 & 54.2 & 58.7 & 60.6\\
ARN + \MethodShort & S & \textbf{67.7} & \textbf{74.2} & \textbf{79.6} & \textbf{81.1} & \textbf{84.4} & \textbf{47.3} & \textbf{51.7} & \textbf{55.5} & \textbf{60.1} & \textbf{61.8}\\
\hline
TRX \cite{trx} & S & 77.5 & 88.8 & 92.8 & 94.7 & 96.1 & 50.5 & 62.7 & 66.9 & 73.5 & 75.6 \\
TRX + \MethodShort & S & \textbf{79.2} & \textbf{89.2} & \textbf{93.2} & \textbf{95.0} & \textbf{96.3} & \textbf{51.9} & \textbf{63.8} & \textbf{68.2} & \textbf{74.4} & \textbf{77.0} \\
\hline
C3D-PN \cite{pn} & T & 50.9 & 61.9 & 67.5 & 72.9 & 75.4 & 28.8 & 38.5 & 43.4 & 46.7 & 49.1\\
C3D-PN + \MethodShort & T & \textbf{52.5} & \textbf{63.8} & \textbf{70.1} & \textbf{75.2} & \textbf{78.2} & \textbf{29.9} & \textbf{40.1} & \textbf{44.5} & \textbf{47.7} & {50.8}  \\
\hline 
ARN \cite{arn} & T & 61.2 & 70.7 & 75.2 & 78.8 & 80.2 & 31.9 & 42.3 & 46.5 & 49.8 & 53.2\\
ARN + \MethodShort & T & \textbf{63.9} & \textbf{73.1} & \textbf{77.4} & \textbf{80.4} & \textbf{81.3} & \textbf{33.6} & \textbf{43.7} & \textbf{48.0} & \textbf{51.1} & \textbf{53.8} \\
\hline
TRX \cite{trx} & T & 75.2 & 88.1 & 91.5 & 93.1 & 93.5 & 33.5 & 46.7 & 49.8 & 57.9 & 61.5\\
TRX + \MethodShort & T & \textbf{76.8} & \textbf{88.9} & \textbf{92.7} & \textbf{93.8} & \textbf{94.1} & \textbf{35.0} & \textbf{48.1} & \textbf{51.1} & \textbf{59.2} & \textbf{62.1}\\
\end{tabular}}
\caption{Results on UCF101 for the Few-Shot Learning setting, with different splits. Accuracies are reported for 5-way, 1, 2, 3, 4, 5-shot classification. S corresponds to the split used in \cite{arn,trx} and T is the TruZe split \cite{truze}, which avoids overlapping classes with Kinetics.}
\label{tab:fslucf}
\end{table}

\subsection{Augmenting the Full Training Set}

We finally explore the effect of augmenting the full dataset, both for smaller datasets, and the large-scale Kinetics. 
Results can be found on Table~\ref{table:fullset}. Again, \Method improves the performance on all datasets even for a pre-trained model. 

\begin{table}[t]
\centering

\begin{tabular}{l l l c}
Augmentation & Dataset & Pretrained & Top-1 \\
\hline 
Standard & UCF101 & No Pretraining & 55.7 \\
ActorCut \cite{actorcut} & UCF101 & No Pretraining & 68.3 \\
\MethodShort & UCF101 & No Pretraining & {\bf 73.1} \\
\hline 
Standard & HMDB51 & No Pretraining & 40.8 \\
ActorCut \cite{actorcut} & HMDB51 & No Pretraining & 44.5 \\
\MethodShort & HMDB51 & No
Pretraining & {\bf 46.4} \\
\hline 
Standard & UCF101 & Sports1M &  93.6 \\
\MethodShort & UCF101 & Sports1M & {\bf 95.3} \\
\hline 
Standard & HMDB51 & Sports1M & 66.6 \\
\MethodShort & HMDB51 & Sports1M &  {\bf 68.4} \\
\hline
Standard & Kinetics & Sports1M & 75.4 \\
\MethodShort & Kinetics & Sports1M &  {\bf 76.3}\\
\end{tabular}
\caption{Augmenting standard datasets improves classification even with a model pre-trained on the largest existing dataset (Sports1M). %In particular, we see an increase of 17.4\% when training UCF101 from scratch .
}
\label{table:fullset}
\vspace{-0.5cm}
\end{table}

\section{Why Not Intra-class Augmentation?}
\label{intra-class}
One other possibility we explored is intra-class augmentation instead of using semantic classes. However, when we followed the same procedure on 20\% labeled data of UCF101 we obtain an accuracy of 41.4\% in comparison to 58.9\% when using semantically similar classes. Similarly, in Kinetics100 we obtain an accuracy of 50.1\% and 54.4\% using 5\% and 10\% labeled data respectively. That is 9.4\% and 8.9\% lower than the results using semantic neighbors. We believe there to be two main concerns in intra-class augmentation. The first is that Cutmix \cite{yun2019cutmix} has been shown to be an excellent regularization technique. This is aided by having samples that have soft labels (since they are a ratio of samples from different classes). However, using intra-class augmentation would force the labels to be the same as the ground truth class. The second reason is that samples of a particular class are clips that were part of the same video.
%\marcus{one could prevent that? we should do such an experiment, at least before arxiv} 
This is the case in both HMDB51 and UCF101 and not so in Kinetics100. If we cut the background from one sample and paste the foreground onto this, it results in an identical sample to the original foreground sample. This is because the background is the same in both cases. All we end up doing then is training the model on multiple instances of the same data which leads to overfitting and hence a poor accuracy at test time. However, since the results are much worse for Kinetics100 as well, we believe that this could be a smaller contributing factor.

\section{Distillation Loss for Semi-Supervised Learning (SSL)}
\label{dist-loss}
Given frame $a$ from video $v$, to distill appearance information of objects of interest, we use the softmax predictions of a ResNet \cite{resnet} image classifier. This network is pre-trained on Imagenet and not modified during training. Let the output of the ResNet be denoted as $h(a) \in \mathbb{R}^{M}$ where $M$ = 1000 which is the number of classes in Imagenet.  We randomly select a frame from all videos (labeled, unlabeled and augmented) for training. %\marcus{the following sentences are a bit unclear to me} The classifier model \marcus{not clear what classifier model you refer to.} 
The classifier model in our architecture, produces an embedding $q(v) \in \mathbb{R}^{M}$ which is of the same dimensions and space of $h(a)$.
%\marcus{are you sure it is the same space, as later on you use log in Eq. 1.}. 
We train $q(v)$ to match the output of $h(a)$ by using a soft cross-entropy loss that treats the ResNet outputs as soft labels. This loss $\mathcal{L}_{d}$ can be seen in Eq.~\ref{eq:dist}. Our final loss function is a combination of $\mathcal{L}_{d}$ and $\mathcal{L}_{s}$ (categorical cross-entropy loss for video samples). This is done following the work in VideoSSL \cite{Jing_2021_WACV}.

\begin{equation}
 \mathcal{L}_{d}=-\sum_{v\epsilon (X\cup Z), a\epsilon v}^{}h(a)\log \left( q(v) \right)   
\label{eq:dist}
\end{equation}

\section{Analysis of Number of Augmented Samples}
\label{number_of_samples}
\begin{figure}[]
    \centering
    \includegraphics[width=0.45\textwidth]{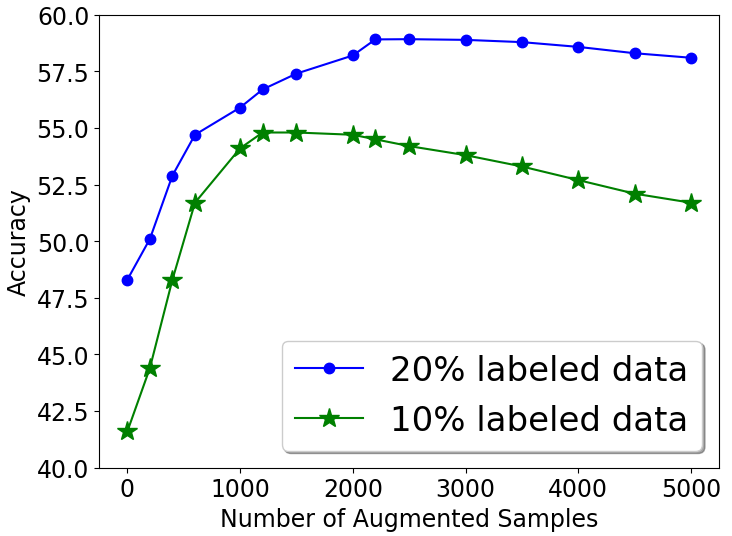}
    %\hspace{0.5cm}
    %\includegraphics[width=0.1\textwidth]{./figures/legend.png} 
    \caption{Comparison of performance with increasing number of augmented samples. Results are for 10\% and 20\% of labeled data UCF101. We see that the performance increases initially, reaches a peak and slowly starts dropping.
    }
    \label{fig:comparison}
\end{figure}

We see a common pattern when adding augmented samples to the different SSL settings. This basically refers to increasing the number of augmented samples in the training set.
%\marcus{are you talking about increasing the number of examples?}. 
We see that the accuracy increases initially, reaches a peak performance and then starts dropping slowly as can be seen in Figure~\ref{fig:comparison}.
%\marcus{do you have a figure for that?}. 
This makes sense as we don't expect every mixed example to be helpful for training. In fact, this helps us to define $\omega_{i}$ for the selector.
%\marcus{This is confusing, I think, you seem to be talking mainly about the number of examples, but $\omega$ is also/mainly about the quality of the examples}. 
We can see Figure~\ref{fig:comparison} for the results from 0 augmentations to 5000 for 10\% and 20\% labeled data on UCF101. The sweet spot for the 10\% labeled data is around 1200 augmentations and for the 20\% labeled data is around 2000 augmentations. Both of which are obtained using $\omega_{i} = 0.6$. We decide the value of $\omega_{i}$ based on these and results and use the same for HMDB51 and Kinetics100 for all settings. If we increase the value of $\omega_{i}$ we obtain fewer samples and decreasing the value of $\omega_{i}$ results in more number of samples for training. The value of $\omega_{i}$ thus determines the number of augmented samples and also their quality.

\section{Other Selector Choices}
\label{selector}

The design of the selector is a crucial aspect of our model. We want the selector to be able to learn what makes a good pair of videos for mixing without actually having to mix every single pair.
%\marcus{but the following seems more a comparison to related work}. 
 However, for lower percentages of labeled data, we can generate all possible samples of semantic
classes and convert a state-of-the art frame selection model
(SMART) \cite{smart} to do sample importance instead of frame importance.
We also consider a simple baseline of using a discriminator network to pick only realistic samples. We report the results in Table~\ref{tab:selector}. Another approach was to randomly pick a certain amount of samples to train the classifier network. %We see that our proposed selector does as well as DVL \marcus{what is that?} whilst only observing a fraction of the samples \marcus{not in the table but should be!}. 

We not only outperform all alternative approaches, we also do this by saving on both memory and computation cost. For example, in the 20 percent setting, SMART sees
99K videos and these 99k videos have to be precomputed
and stored before training SMART. However, the proposed
approach only needs 12K videos and outperforms SMART
by up to 1.4\%. This analysis is only to show a comparison to possible alternatives when storing data is feasible. The idea of trying these alternatives is only
feasible in low percentage labeled data of small datasets like
UCF101 and HMDB51. Even 50\% labeled data in UCF101, results in having to mix over 400k videos while large scale datasets like Kinetics400 would lead to millions of mixes being needed making it practically unfeasible.

\begin{table}[htb]
\begin{center}

\begin{tabular}{lllllllll}
              & \multicolumn{2}{c}{50\%}     & \multicolumn{2}{c}{20\%}     & \multicolumn{2}{c}{10\%}     & \multicolumn{2}{c}{5\%}      \\
Method        & Acc     & SS & Acc     & SS & Acc      & SS & Acc     & SS \\ 
\hline
Random        & 61.9          & 430K         & 56.2         & 99K          & 51.8          & 44K          & 42.3          & 9.7K         \\
Discriminator & 62.8          & 430K         & 57.3         & 99K          & 52.2          & 44K          & 41.1          & 9.7K         \\
SMART \cite{smart} & 68.9 & 430K & 58.9 & 99K & 57.8 & 44K & 46.5 & 9.7K \\
Proposed      & \textbf{72.1} & 39K          & \textbf{60.3} & 12K          & \textbf{56.1} & 5.2K         & \textbf{48.0} & 1.2K     \\   
\end{tabular}
\end{center}
\caption{Comparison of approaches for the use of Selector. All results are reported on UCF101. 'Acc' corresponds to accuracy and 'SS' corresponds to the number of mixed videos that the Selector looks at. All results are on different percentage of labeled data in UCF101.}
\label{tab:selector}
\end{table}

\section{Why Re-train the Classifier Network?}
\label{re-train}
%\marcus{is it clear for the reader what the "classifier network" is referring to?}
Here, we are talking about the classifier network in our proposed architecture that the selector learns from (based on the validation loss).
Training the Selector and the Classifier together is also possible. But we decide against this for 2 reasons. First, and the most important reason is that we want to save out on computational cost needed to generate an augmented sample. We showed that the selector network looks only at a fraction of samples before it understands what makes a good pair. Hence, we first train the selector by generating augmented samples taken from random samples of semantically similar classes. Once the selector is trained, we don't need to generate the mixed sample for all possible pairs and only generate the mixed samples for good pairs (the selector need not have seen these pairs before). We then augment the original dataset by samples that the selector believes will improve the classifiers performance %\marcus{I think you would have to give some numbers here, to be convincing. Because you already have a trained classifier at the point you finish the selector training, so one would need 0 additional examples to train the classifier}. \marcus{even if you used the classifier network, you could still transfer the selector, no?} \marcus{Can you say what is better? the jointly trained vs. the retrained}
We compare the performance of the joint training and re-training of the classifier network in Table~\ref{tab:retrain}. We see that re-training the classifier network always yields the best performance.

\begin{table}[htb]
\begin{center}
\begin{tabular}{lllll}
Method        & 50\% & 20\% & 10\% & 5\%  \\
\hline
Jointly trained & 66.5 & 57.4 & 53.1 & 44.7 \\
Retrained  & \textbf{72.1} & \textbf{60.3} & \textbf{56.1} & \textbf{48.0} \\
\end{tabular}
\end{center}
\caption{Comparison of jointly training classifier and re-training it. We see that there is a consistent large improvement in re-training the classifier.}
\label{tab:retrain}
\end{table}

\section{Examples of Selected and Discarded Samples}
\label{examples}

\begin{figure*}[]
    \centering
    \includegraphics[width=\textwidth]{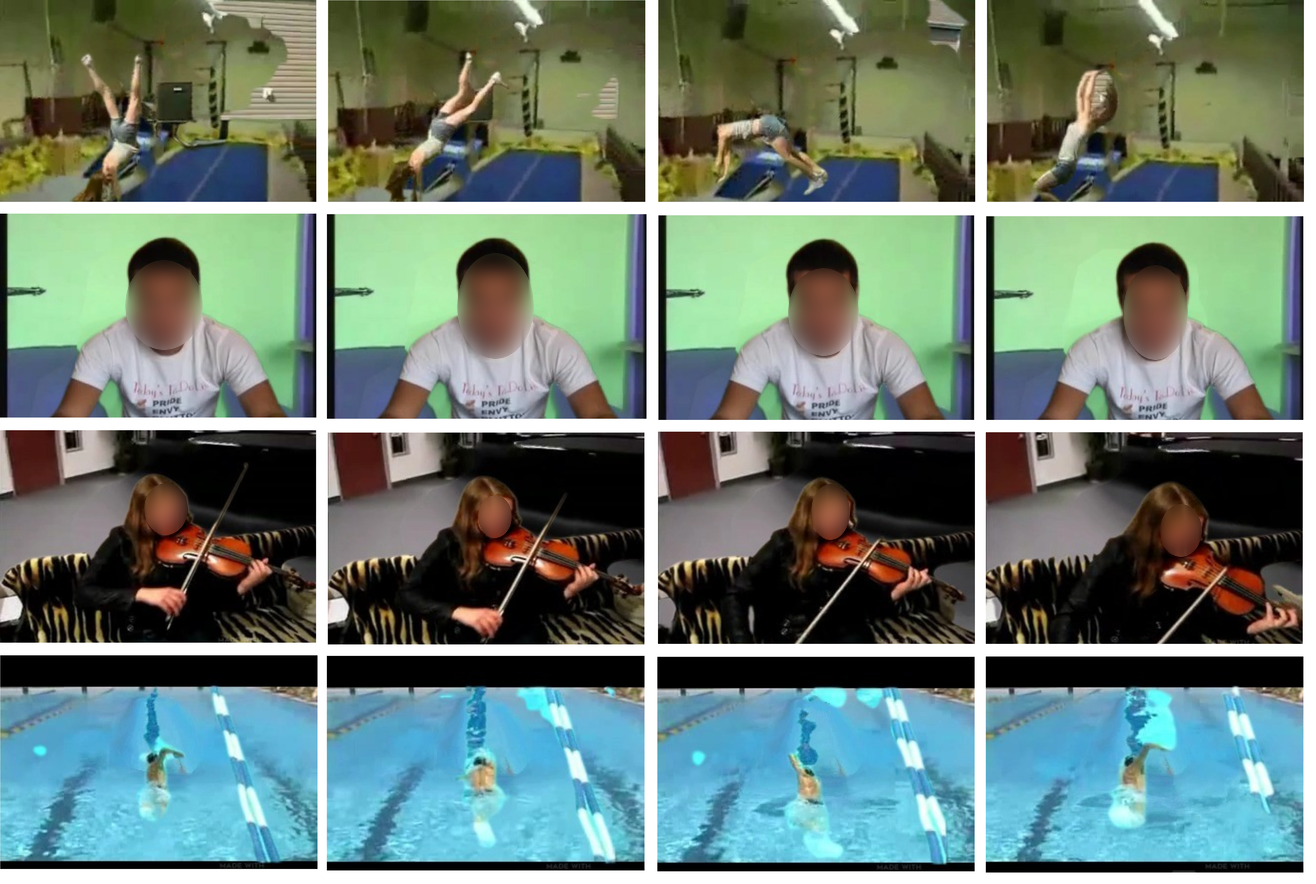}
    %\hspace{0.5cm}
    %\includegraphics[width=0.1\textwidth]{./figures/legend.png} 
    \caption{Visualizing selected examples. From top to bottom as (foreground, background) pairs: (flic-flac, cartwheel), (smile, laugh), (playing violin, playing cello), (front crawl, swimming backstroke). The first two are examples from HMDB51 and the last two from UCF101.
    }
    \label{fig:select}
\end{figure*}

To understand what made a good sample we visualize a few samples that were selected by the selector model and a few samples that were discarded. These can be seen in Figure~\ref{fig:select}. The samples are displayed as 4 frames for better visualization. Based on the small subset of examples seen, we believe that for good pairs to be selected some of the criteria could be coherent inpainting, similar camera movement, not too drastic a background change.

\begin{figure*}[]
    \centering
    \includegraphics[width=\textwidth]{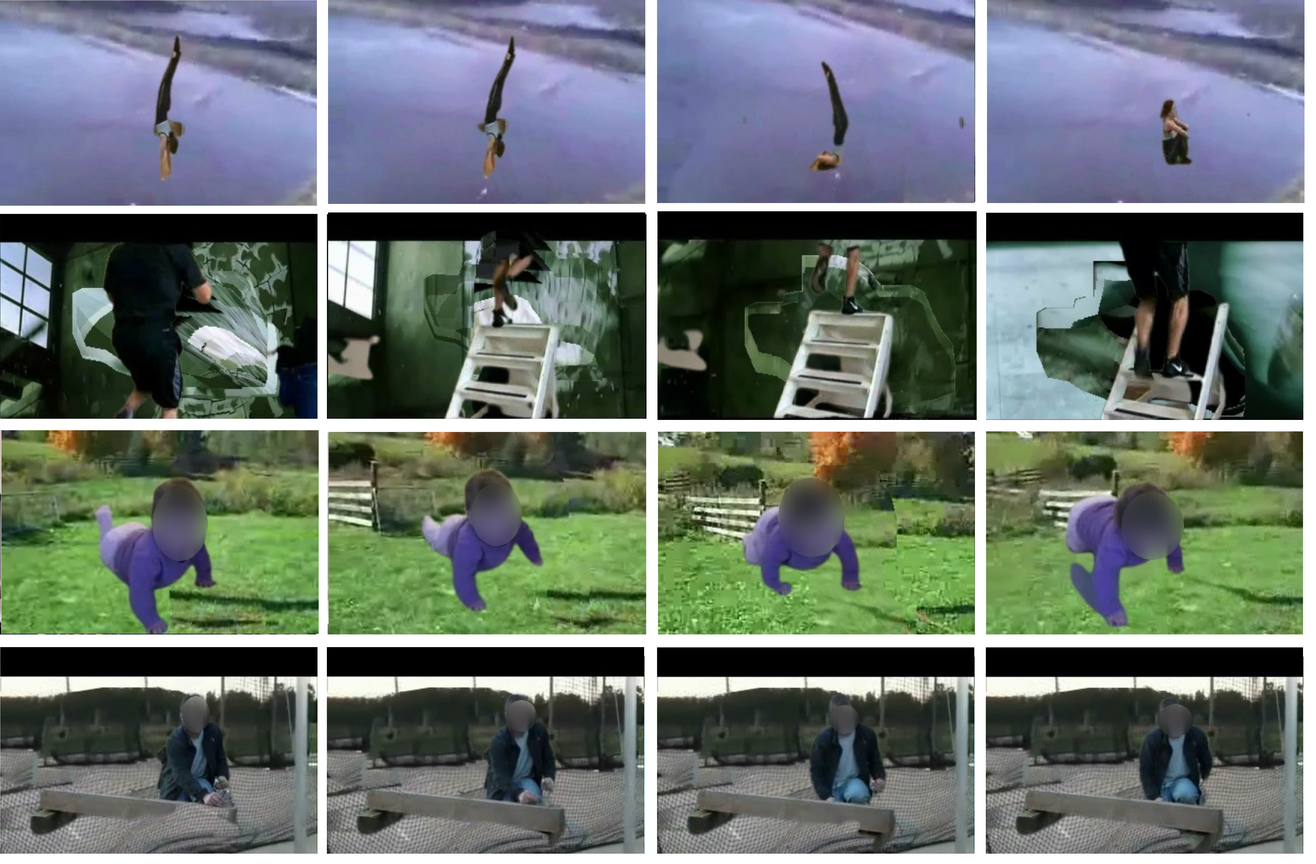}
    %\hspace{0.5cm}
    %\includegraphics[width=0.1\textwidth]{./figures/legend.png} 
    \caption{Visualizing discarded examples. From top to bottom as (foreground, background) pairs: (somersault, diving), (climbing stairs, falling floor), (baby crawling, walking dog), (hammering, hammer throw). 
    }
    \label{fig:discarded}
\end{figure*}

We see some samples of discarded examples in Figure~\ref{fig:discarded}. Based on the small subset of examples seen we think possible bad pairs are due to bad video compositing (example 2 in Figure~\ref{fig:discarded}), varying camera movements (example 3 in Figure~\ref{fig:discarded}) or a drastic change in background (example 1 in Figure~\ref{fig:discarded}). These are however based on the few examples we see.

\section{Effect of Semantic Match in generalization ability.}
\label{gen_ability}
 We test the generalization ability of the semantic matching by comparing it with random matching which would correspond to row 4 of Table 1 in the main paper. We observe that the performance does decrease. To strengthen this test, we tried the same experiment in the FSL setting, which is an extreme case for generalization. We augment data for two different methods, using the proposed L2A, using both semantic and random matching of classes. We observe that even in this setting, which is the most susceptible to overfitting, the semantic matching outperforms random matching. We will add this to the final version. %, and report results below using random matching and semantic matching for 3 different settings on HMDB51 using the TruZe split.

\begin{table}[htb]
\centering
\resizebox{0.6\columnwidth}{!}{\begin{tabular}{|l|l|l|l|l|}
\hline
Method                        & Class Matching & 1-shot & 3-shot & 5-shot \\ \hline
C3D-PN & Random   &    28.1    &   42.9     &    47.7    \\ \hline
C3D-PN & Semantic &   29.9 &   44.5 &   50.8 \\ \hline
TRX & Random       &   33.5     &   49.9     &   60.3     \\ \hline
TRX & Semantic &    35.0 &  51.1 &  62.1   \\ \hline
\end{tabular}}
\caption{Results on FSL using the proposed Semantic Matching vs random matching using the TruZe \cite{truze} split.}
\end{table}

\section{Limitations and Future Work}

The main area of improvement is the time needed for training. Optimizing the Selector with RL is time-consuming, and so is compositing the initial samples for training it. Future work could address this by parameterizing the composition process and learn these parameters instead of compositing the pairs directly. It could also learn to select particular frames in a video, and avoid the computational cost of temporal redundancy. Finally, another possible direction is to learn what samples to discard from the initial dataset itself.

% The main limitation for the approach is the time needed for training. The selection network optimization with REINFORCE adds a time overhead in comparison to standard augmentation. Along with this, the \VideoMixShort adds an overhead (155.1 GFlops for 20\% labeled data in UCF101). This cost, however, is only incurred during training. As possible future work, we could parameterize the composition process and learn these parameters instead of compositing the pairs directly. This would reduce compute time. Another possible direction is to learn what samples we could discard from the main dataset itself.

%\ls{parameterize composition process and learn those} 

%\ls{learn to discard samples of the main dataset itself}

\section{Conclusion}

While standard data augmentation strategies in action recognition are hand-crafted, we propose to learn %good augmented samples by learning 
which pairs of videos are good to composite. In order to do this, our approach leverages three components. We train a Selector optimized with RL to choose which pairs of videos are good to composite. We reduce the search space by using samples from semantically similar classes. We perform a clean segmentation for mixing samples and remove actors as well as objects from foreground and background samples. With this, we obtain state-of-the-art results in semi-supervised and few-shot action recognition settings, and improve in the fully supervised setting. In particular, we see gains of up to 8.6\% and 3.7\% in the semi-supervised and few-shot settings. We also see an improvement of up to 17.4\% when compared to standard augmentation in the fully supervised setting when training from scratch.
\bibliographystyle{splncs04}
\bibliography{eccv}
\end{document}